\documentclass[letterpaper, 10 pt, conference]{ieeeconf}  % Comment this line out if you need a4paper

\usepackage{graphicx}
\IEEEoverridecommandlockouts                              % This command is only needed if 
\usepackage{soul, color}
\usepackage{url}
\usepackage{amsfonts}
\usepackage{siunitx}
\usepackage{algorithm}
\usepackage{algpseudocode}
\usepackage{amsmath}
\usepackage{svg}
\usepackage{hyperref}
\usepackage{pifont}% http://ctan.org/pkg/pifont
\usepackage{wrapfig}
\usepackage{subcaption}

\usepackage{graphicx, caption, subcaption}
\usepackage{array, multirow}
\usepackage{xcolor}
\graphicspath{ {figures/} }

\newcommand{\vect}[1]{\boldsymbol{#1}}
\DeclareMathOperator*{\argmin}{argmin}

\newcommand{\cmark}{\ding{51}}%
\newcommand{\xmark}{\ding{55}}%

\title{\LARGE \bf
Kinematic Nonlinear Spatio-Temporal Trajectory Warping for Contact-Rich Dexterous Manipulation Demonstrations
}

\author{Hyojae Park$^{1}$, Arjun S. Lakshmipathy$^{1}$, Nancy S. Pollard$^{1, 2}$
\thanks{The authors are with the $^{1}$Computer Science Department and the $^{2}$Robotics Institute at Carnegie Mellon University, Pittsburgh, USA.}
\thanks{Corresponding author’s contact: {\tt\small hyojae@cmu.edu}}}

\begin{document}

\maketitle
\thispagestyle{empty}

\global\csname @topnum\endcsname 0
\global\csname @botnum\endcsname 0

%%%%%%%%%%%%%%%%%%%%%%%%%%%%%%%%%%%%%%%%%%%%%%%%%%%%%%%%%%%%%%%%%%%%%%%%%%%%%%%%
\begin{abstract}

We present a straightforward but effective method for repurposing existing contact-rich dexterous manipulation demonstrations. Starting from inputs of hand and object trajectories, our method outputs high-quality nonlinear trajectory warps that account for intermediate waypoints, environmental barriers, temporal shifts, and varied start/end configurations. Foundational to our method is the utilization of contact distributions, which we show allows us to reliably compute complex and high-dimensional dexterous hand trajectories following a simple object-centric warp specification pipeline. We evaluate our method across 12 variations sourced from 4 demonstrations in a publicly available dataset of human hand motion data, perform baseline comparisons, and demonstrate generalization of our approach to different manipulators. Results and code will be made available on publication.

\end{abstract}

\begin{figure}%
    \centering 
{\includegraphics[width=0.99\linewidth]{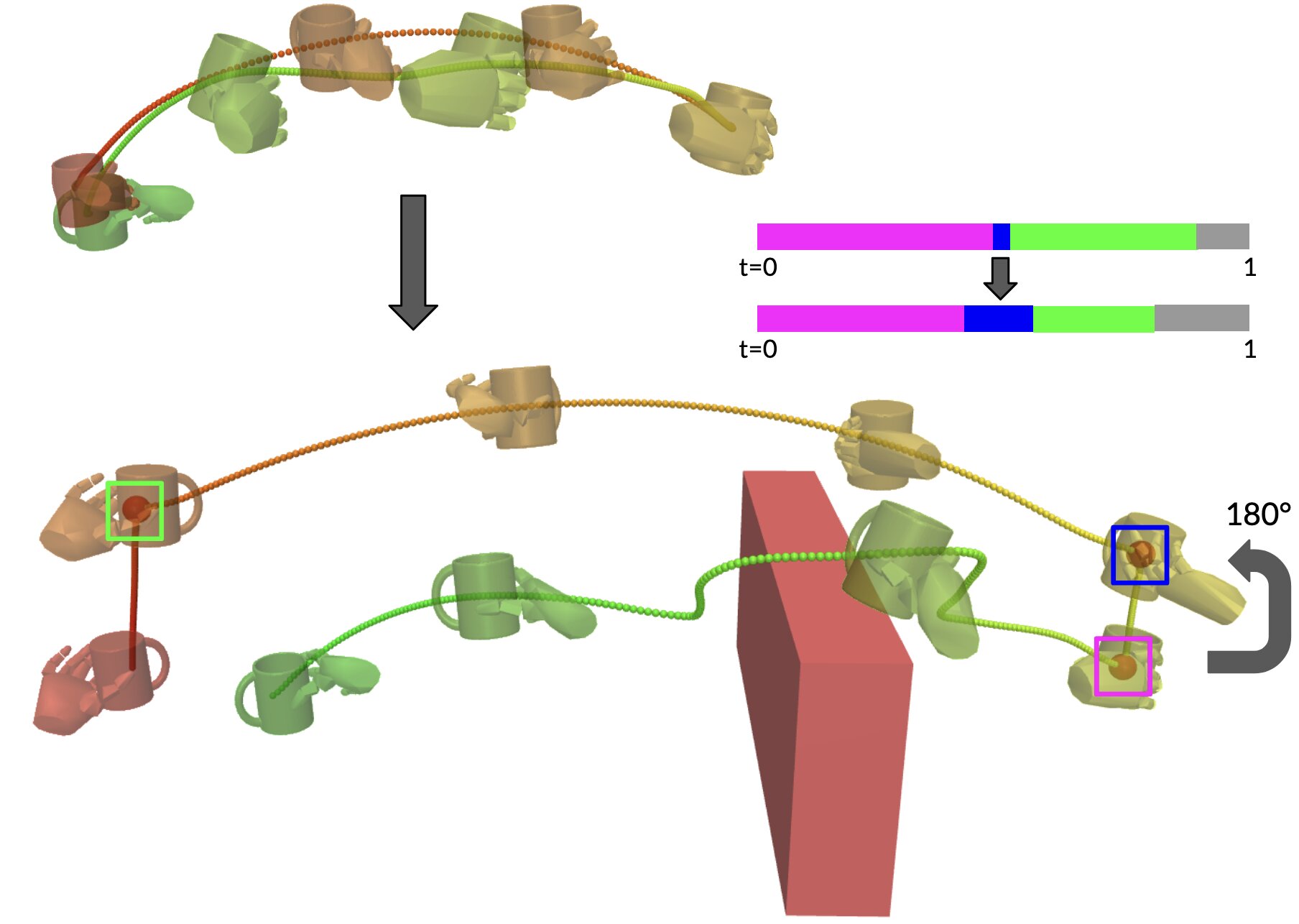}} \captionof{figure}{Our method warps hand and object trajectories from existing contact-rich demonstrations to fit spatial and temporal waypoints, environmental barriers, and new start/end positions. Approximate mappings between the original and warped trajectories are color-coded for visualization.}
\label{fig:main}
\vspace{-0.35cm}
\end{figure}

%%%%%%%%%%%%%%%%%%%%%%%%%%%%%%%%%%%%%%%%%%%%%%%%%%%%%%%%%%%%%%%%%%%%%%%%%%%%%%%%

\section{INTRODUCTION}

Despite significant advances in optical~\cite{taheri2020grab,fan2023arctic}, wearable~\cite{mao2025visuo,guzey2025dexterity,yin2025osmo}, and tele-operative~\cite{Handa2019DexPilotVT,sivakumar2022telekinesis,naughton2024respilot,yin2025geometric} motion capture interfaces and processing algorithms, reconstructing high quality dexterous manipulation demonstrations remains a time-consuming and labor-intensive process. This persistent bottleneck in data acquisition continues to hinder learning-from-demonstration pipelines, driving the demand for methods capable of adapting existing demonstrations to new environmental variations. Furthermore, sampling across variations is vital for hardening sim-to-real transfer strategies against real-world uncertainty. Specifying exactly \textit{how} such variations are generated, as well as what variables to sample over, is a core problem in domain randomization.

Variations can generally be formulated as \textit{nonlinear trajectory warps}, allowing for the modification of spatial constraints (goals, waypoints, obstacles), temporal constraints (timing of targets or segments), and the method of interpolation while attempting to preserve characteristics of the original motion. Manipulation also adds complexity because the hand and object trajectories are implicitly coupled; breaking the coupling can result in serious artifacts such as motion misalignment, penetrations, or loss of contact. Effective nonlinear warping algorithms for dexterous manipulation must account for all these factors simultaneously; however, no simple ``off-the-shelf" solver currently exists for this task.

We address this gap by presenting an effective method for nonlinear trajectory warps of long sequence dexterous manipulation demonstrations. Starting from inputs of hand and object trajectories, our method generates high quality coupled warps that respect arbitrary barriers, spatial and temporal waypoints, and start/end configurations. Our main insight is to split the problem into a two-stage object-centric pipeline. We first warp the object's trajectory and then recover the hand pose using contact distributions, defined as per-frame point correspondences between the hand and object meshes. %that encode the spatial hand-object relationship of the original demonstration.
Furthermore, this contact-based approach ensures that our method remains reliable even under large nonlinear spatio-temporal transformations where naive interpolation-based approaches struggle.

% This allows us to easily decompose the problem into an object-centric pipeline that first performs the warp on the object and subsequently recovers the hand trajectory using contact distributions.

We demonstrate our method's effectiveness across 12 variations sourced from 4 demonstrations in a publicly available dataset of human hand motion data, perform baseline comparisons, and demonstrate generalization of our approach to different manipulators. Our method runs in minutes on a CPU and accommodates substantial warps under a consistent strategy with few parameters beyond input specifications. Our method can be used as a \textit{modular drop-in replacement} in dexterous manipulation data augmentation pipelines, which we will facilitate through public code release.

\section{RELATED WORK}

We examine three categories of related work: data augmentation, contact-driven grasp and manipulation synthesis, and motion planning.

\subsection{Data Augmentation}

Data augmentation techniques enhance a model's robustness and ability to generalize when working with limited original samples. While several techniques focus on image~\cite{jin2025physically,kostrikov2020image} or semantic~\cite{chen2023genaug,bharadhwaj2024roboagent} variation, these techniques do not explicitly alter trajectories; rather, trajectories are either replayed~\cite{jin2025physically} or left to be learned by a model~\cite{chen2023genaug}.

Trajectory altering augmentation techniques directly output computed warped trajectories from an input specification. Leading the field in adoption, the MimicGen line of work~\cite{mandlekar2023mimicgen,jiang2025dexmimicgen} and its variants~\cite{xue2025demogen,oh2025self} currently represent one of the most common choices for trajectory-based data augmentation. These techniques, like ours, adopt an object-centric warping strategy; however, the extent of variation is limited to \textit{rigid, linear} transformations ($SE(3)$). More concretely, these works assume that the configuration of the hand relative to the object is fixed under $SE(3)$ equivariance and that new samples are generated by multiplying per-timestep object configurations by a static transformation $\vect{M} \in SE(3)$. While simple and efficient, these assumptions make it impossible to modify intermediate object paths, adjust timings, or navigate around obstacles. Our method offers generalization to nonlinear warps that accommodate these modifications while maintaining simplicity. Furthermore, our method is specifically designed for dexterous manipulation by optimizing the full hand articulation per timestep rather than applying fixed end-effector transforms \cite{jiang2025dexmimicgen} and requires only minor additional computational overhead.

\subsection{Contact-Driven Grasp and Manipulation Synthesis}

Contacts have played a vital role in grasp and manipulation synthesis techniques. They have frequently been leveraged as a loss term for pose optimization~\cite{lakshmipathy2023contactedit,turpin2022grasp,brahmbhatt2019contactgrasp,ye2012synthesis,hazard2020automated}, a constraint for training generative models~\cite{christen2022d,wu2022saga}, a prior for retargeting human trajectories~\cite{lakshmipathy2025kinematic,mandi2025dexmachina,pan2025spider}, and a primitive for motion planning~\cite{pang2023global,cheng2022contact}. Modern high quality human motion and interaction datasets~\cite{ taheri2020grab,fan2023arctic,delpreto2022actionsense,song2025opentouch}, many of which serve as the backbones and testbeds of manipulation research efforts, are notably prioritizing the inclusion of detailed contact information. Our work expands this literature by using contact to maintain motion coupling during non-trivial spatio-temporal trajectory warps.

\subsection{Motion Planning}

Motion planning strives to find a sequence of configurations (kinematics) or states/actions (dynamics) that satisfy terminal or intermediate constraints. These approaches span a range of paradigms including offline collision-free trajectory optimization~\cite{ratliff2009chomp, kalakrishnan2011stomp}, contact sequence planning~\cite{cheng2022contact, pang2023global, suh2025dexterous, le2024fast}, and learning-based approaches~\cite{carvalho2023motion, seo2025presto,dalal2024neural, qureshi2019motion}. These techniques can be broadly grouped into local, global, sampling-based, or gradient-based methods~\cite{zhang2025motion}, and are typically designed to compute solutions from \textit{scratch} or under \textit{sparse initial conditions}.

While sharing similar input-output profiles, motion planning and trajectory warping employ divergent methodologies: the former constitutes a ``bottom-up" constructive approach, while the latter functions as a ``top-down" refinement of existing data. Consequently, motion planning often struggles to produce trajectories with nuanced characteristics (e.g., smooth, natural movements) while trajectory warping faces the inverse challenge of preserving these characteristics while adapting to new task goals. Our method lies within the warping paradigm: it adapts the trajectory to new spatio-temporal constraints while attempting to preserve features of the original motion.

\section{METHOD}
\vspace{-7pt}
\begin{figure}[h]%
    \centering
    % \subfloat[\centering object trajectory]{{\includegraphics[width=8.7cm]{figures/overview_object.png} }}%
    % \qquad
    % \subfloat[\centering hand trajectory]{{\includegraphics[width=8.7cm]{figures/overview_hand.png} }}%
    {\includegraphics[width=\linewidth]{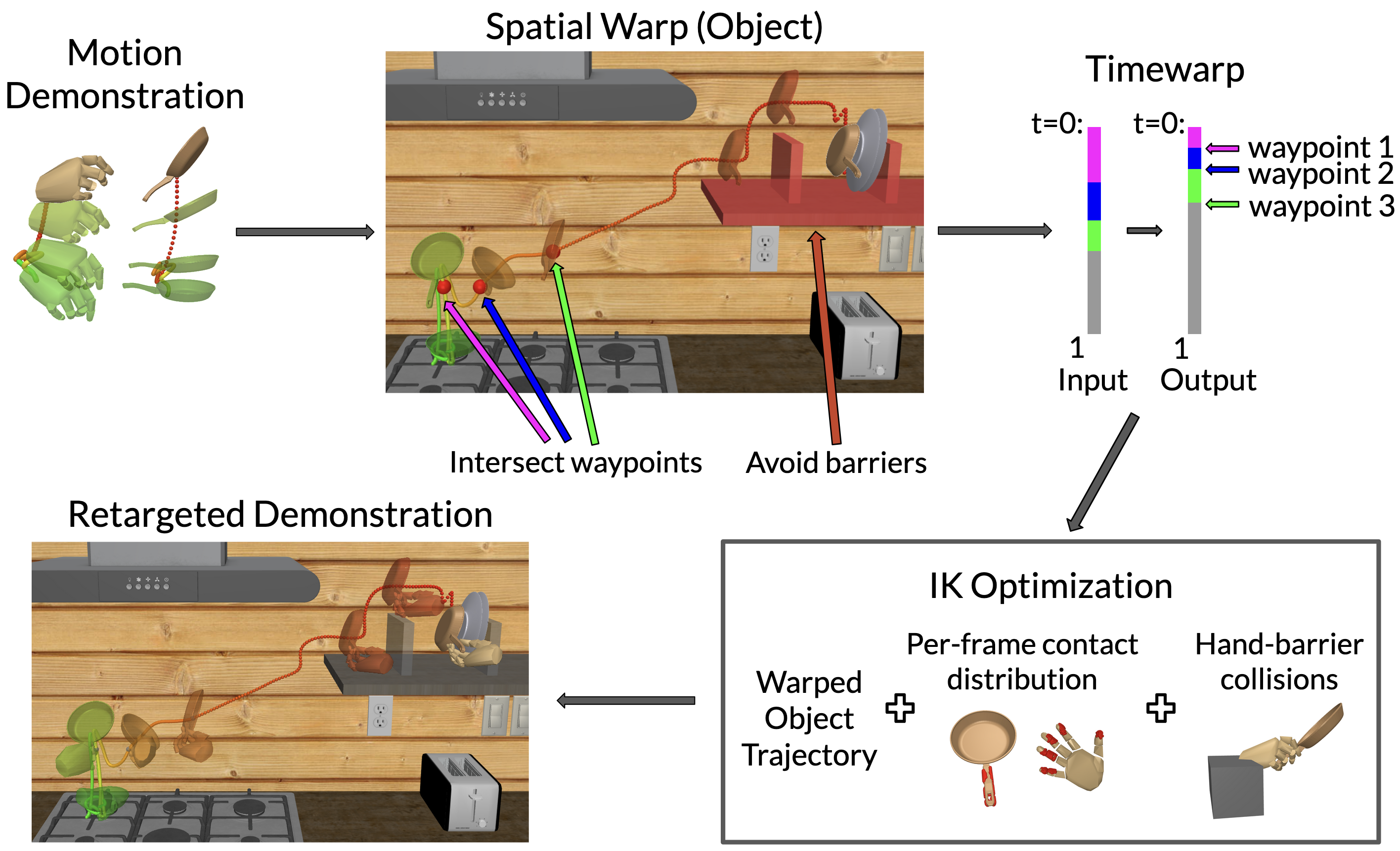}}
    \caption{%Overview of our pipeline. Starting from inputs of an existing manipulation demonstration with contact data, spatio-temporal waypoints, barriers, and start/end configurations, our method (a) first warps the object trajectory to satisfy the inputs and then (b) computes the corresponding hand trajectory using hand-object contact distributions.
    System overview. Given a motion demonstration, our method first spatially warps the object trajectory to intersect user-specified waypoints and avoid scene barriers. It then applies a time-warp to ensure the trajectory satisfies the time constraints of each waypoint. Finally, an inverse kinematics (IK) optimization recovers the hand motion by tracking the warped object trajectory, subject to hand–barrier collision constraints, using per-frame corresponding contact distributions to yield the complete retargeted demonstration.}%
    \label{fig:overview}%
\end{figure}

Figure \ref{fig:overview} illustrates our object-centric warping pipeline. Starting from inputs of polyline-represented trajectories of the object ($\mathcal{T}_o$) and hand ($\mathcal{T}_h$), corresponding hand-object contact distributions per frame, spatio-temporal waypoints, barriers, and terminal configurations $p_{\text{start}}'$ and $p_{\text{end}}'$, our pipeline constructs the final warp in four stages. We begin by spatially warping the object trajectory so that it passes through the specified positions and orientations. Next, we further warp this trajectory to prevent object intersections with a set of barrier meshes. Third, we fit a B-spline to the resulting object trajectory to respect temporal constraints. Finally, we utilize hand-object contact distributions to compute the corresponding warped hand trajectory per frame via IK optimization while accounting for barrier constraints.

Each waypoint is defined as $w=(p, r, t, t')$. The positional constraint $p\in \mathbb{R}^3$ specifies a point in world space through which the object must pass. The rotational constraint $r$ defines an imposed rotation applied over the segment between the previous and current waypoint, taking effect when the trajectory reaches $p$. The temporal constraint is defined by two parameters $t$ and $t'$, where $t$ is the timestep in the original trajectory at which the object must pass through $p$ and $t'$ is the corresponding timestep in the output trajectory. Each barrier is a mesh in world space representing a region that the trajectory must avoid. $p_{\text{start}}'$ and $p_{\text{end}}'$ alter the starting and ending position of the object, respectively, and are implicitly defined as the first and last waypoints. %We provide more details in the proceeding subsections.

\subsection{Spatial Waypoint-Constrained Object Warping}

In this first stage, we use $p_{\text{start}}'$, $p_{\text{end}}'$ and a tuple of waypoints $\mathcal{W} = {w_1,...,w_k}$ to warp the object trajectory. We assume that all trajectories consist of pre-, in-, and post-contact segments. If $p_{\text{start}}'$ and $p_{\text{end}}'$ are not provided, warping is strictly confined to the in-contact segment of the trajectory, which we assume to be a contiguous. If $p_{\text{start}}'$ or $p_{\text{end}}'$ are provided, the pre- and post-contact segments are rigidly shifted to the new boundary positions.

Waypoints partition the in-contact sub-trajectory into consecutive segments, which are independently transformed using algorithm~\ref{alg:transform_segment} to satisfy the positional and rotational constraints at their endpoints. Specifically, algorithm~\ref{alg:transform_segment} transforms each segment so that the trajectory passes through $p_i$ at $t_i$ and $p_{i+1}$ at $t_{i+1}$. The algorithm first computes the endpoint displacements $\Delta_s = p_i - \mathcal{T}_o(t_i)$ and $\Delta_e = p_{i+1} - \mathcal{T}_o(t_{i+1})$, then shifts all intermediate points $\mathcal{T}_o|_{(t_i,t_{i+1})}$ by a blended displacement that smoothly transitions from $\Delta_s$ to $\Delta_e$.

The interpolation scheme blends two strategies: a quadratic blending $\Delta_{\text{quad}}$ that preserves the original trajectory shape, and a linear blending $\Delta_{\text{lin}}$ that produces a more direct path between waypoints. The balance between these two interpolations is controlled by a blend factor $\beta \in [0, 1]$, derived from the maximum endpoint displacement threshold $\alpha$ (default = 0.2 m). When the waypoints are close to the original trajectory ($\beta \approx 0$) quadratic blending dominates, preserving the original motion. When the waypoints are far from the original trajectory ($\beta \approx 1$), linear blending dominates, producing a more direct transition. This adaptive blending produces more natural movements, as opposed to purely using one blending approach.

Lastly, the algorithm imposes a cumulative rotation $R_\Delta$ at each waypoint using SLERP interpolation to ensure smooth orientation transitions between segments. Note that $t'$ is not used in this step; temporal retiming is deferred until after barrier constraints are resolved.

\begin{algorithm}[b]
\caption{\textsc{TransformSegment}}
\label{alg:transform_segment}
\begin{algorithmic}[1]

\Require 
Trajectory segment 
$\mathcal{T}_o|_{[t_i,t_{i+1}]}$, 
target positions $p_i,\; p_{i+1}$, 
target rotation $r_{i+1}$, 
distance threshold $\alpha$

\Ensure 
Warped segment 
$\mathcal{T}_o'|_{[t_i,t_{i+1}]}$

\State 
$\mathbf{x}(t),\, R(t) 
\gets \textsc{Decompose}\!\big(\mathcal{T}_o(t)\big)$

% \Statex
% \Comment{Endpoint displacements and blend factor}

\State 
$\Delta_s \gets p_i - \mathbf{x}(t_i)$, \quad
$\Delta_e \gets p_{i+1} - \mathbf{x}(t_{i+1})$

\State 
$\beta \gets 
\textsc{Clip}\!\left(
\dfrac{\max(\lVert \Delta_s \rVert,\,\lVert \Delta_e \rVert)}{\alpha},\,0,\,1
\right)^{1/2}$

\State 
$R_{\Delta} \gets \prod_{j=0}^{i+1} r_{j}$ \Comment{Cumulative rotation}

\ForAll{$t \in [t_i,\, t_{i+1}]$}
\Comment{Warp each timestep}

    \State 
    $\hat{t} \gets 
    (t - t_i)\,/\,(t_{i+1}-t_i)$

    \State 
    $\Delta_{\text{quad}} 
    \gets 
    (1-\hat{t})^2\,\Delta_s
    + 
    \hat{t}^2\,\Delta_e$

    \State 
    $\Delta_{\text{lin}} 
    \gets 
    (1-\hat{t})\,\Delta_s
    + 
    \hat{t}\,\Delta_e$

    \State 
    $\mathbf{x}'(t) 
    \gets 
    \mathbf{x}(t) + (1-\beta)\,\Delta_{\text{quad}}
    + 
    \beta\,\Delta_{\text{lin}}$

    \State 
    $R'(t) \gets 
    \textsc{Slerp}
    \big(I,\; R_{\Delta},\; \hat{t}\big)
    \, R(t)$

\EndFor

\State \Return 
$\mathcal{T}_o'|_{[t_i,t_{i+1}]} \gets 
\big(\mathbf{x}'(t),\, R'(t)\big)$

\end{algorithmic}
\end{algorithm}

\subsection{Barrier-Constrained Object Warping}

Next, we warp the trajectory to avoid object intersection with a set of barrier meshes $\{\mathcal{M}_i\}$. We first compute a bounding ball $\mathcal{B}_r$ with radius $r$ for the object. We center $\mathcal{B}_r$ at the origin of the object's frame and compute $r$ by taking the distance to its furthest vertex, which effectively approximates the object's workspace under any relative transform $T \in SE(3)$. For each barrier, $\mathcal{M}_i$, we compute an inflated volume $\mathcal{M}_i'$ that conservatively approximates the Minkowski sum of its convex hull with $\mathcal{B}_r$, i.e., $\textsc{conv}(\mathcal{M}_i) \oplus \mathcal{B}_r$ \cite{de2008computational}. Specifically, we uniformly scale $\textsc{conv}(\mathcal{M}_i)$ about its centroid by a factor derived from $r$, producing a superset of the true Minkowski sum of $\textsc{conv}(\mathcal{M}_i)$. Although this overapproximation may exclude some valid trajectory configurations near barrier surfaces, it is computationally inexpensive and sufficient for resolving object-barrier intersections. Finally, for each trajectory point within $\mathcal{M}_i'$, we project it outward along the ray from the barrier's centroid to the point, stepping incrementally until the point exits the hull surface, resolving all intersections. 

\subsection{Temporal Waypoint-Constrained Object Warping}
\label{sec:finalizing}

Following spatial constraint fitting, we perform a temporal retiming to realize the $t \mapsto t'$ mapping. We first smooth the trajectory by constructing a chord-length parameterized B-spline \cite{piegl2012nurbs} for each segment $\mathcal{T}_o\big|_{[t_i,\, t_{i+1}]}$. This induces a normalized timewarp $\tau_i : [0,1] \rightarrow [0,1]$, which maps uniform time to the spline’s chord-length parameterization. Each local parameterization is then scaled and shifted to the temporal interval $[t_i', t_{i+1}']$, and the per-segment warps are composed into a global timewarp $\tau$. Then, a global cubic B-spline is fit using $\tau$, ensuring that all positional waypoints and temporal constraints are satisfied while $C^2$ continuous motion. Finally, we uniformly sample points from this spline to obtain the re-timed contact segment, which is combined with the unchanged pre- and post-contact segments to produce the final object trajectory $\mathcal{T}_o''$.

\subsection{Hand Warping}

At this point, the object trajectory is fully resolved: it satisfies all spatial, barrier, and temporal constraints. However, the resulting object trajectory $\mathcal{T}_o''$ and timewarp $\tau$ may induce a trajectory that substantially deviates from the original demonstration. Naively recovering the hand trajectory via rigid-body transforms fails under non-uniform timewarps, as we demonstrate in section \ref{sec:baseline}. Instead, we recover the hand trajectory by directly optimizing for contact distributions at every timestep in $\mathcal{T}_o''$.

For each timestep $t$, we retrieve the contact correspondences at $t^*$, the last input timestep mapped by $\tau$ before $t$. This approach is necessitated by the lack of proven methods for interpolating between arbitrary contact distributions; consequently, we retrieve the distributions directly from the original demonstration. We then solve for the hand configuration $\vect{\theta}$ at each output timestep $t$ using the equation:

\begin{equation}
    \vect{\theta}^*(t) = \argmin_{\vect{\theta}} \;
    \frac{\sum_{i=1}^{N(t)} \Gamma_{D,i}(\vect{\theta}, t)}{N(t)} 
    + \sum_b^B \sum_{p \in \mathcal{P}(\vect{\theta}, t)} \Gamma_B(p, b)
    \label{eq:opt}
\end{equation}

where $N(t)$ is the number of contact correspondences at timestep $t$, $B$ is the number of barriers, and $\mathcal{P}(\vect{\theta}, t)$ denotes the hand mesh vertices in world space at timestep $t$ computed via forward kinematics from configuration $\vect{\theta}$. $\Gamma_{D,i}$ penalizes the $L_2$ distance between the $i$-th contact correspondence pair on the hand and object, and $\Gamma_{B}$ penalizes hand-barrier intersections. $\Gamma_{B}$ is formulated as:

$$\Gamma_{B}(p, b) = \left(1 - \frac{\text{SDF}(p, b)}{m}\right)^n$$

\begin{wrapfigure}{r}{0.18\textwidth}
    \centering
    \vspace{-15pt}
    \subfloat[without $\textsc{loss}_\textsc{barrier}$]{
        \includegraphics[trim={24cm 14cm 20cm 4cm}, clip, width=0.92\linewidth]
        {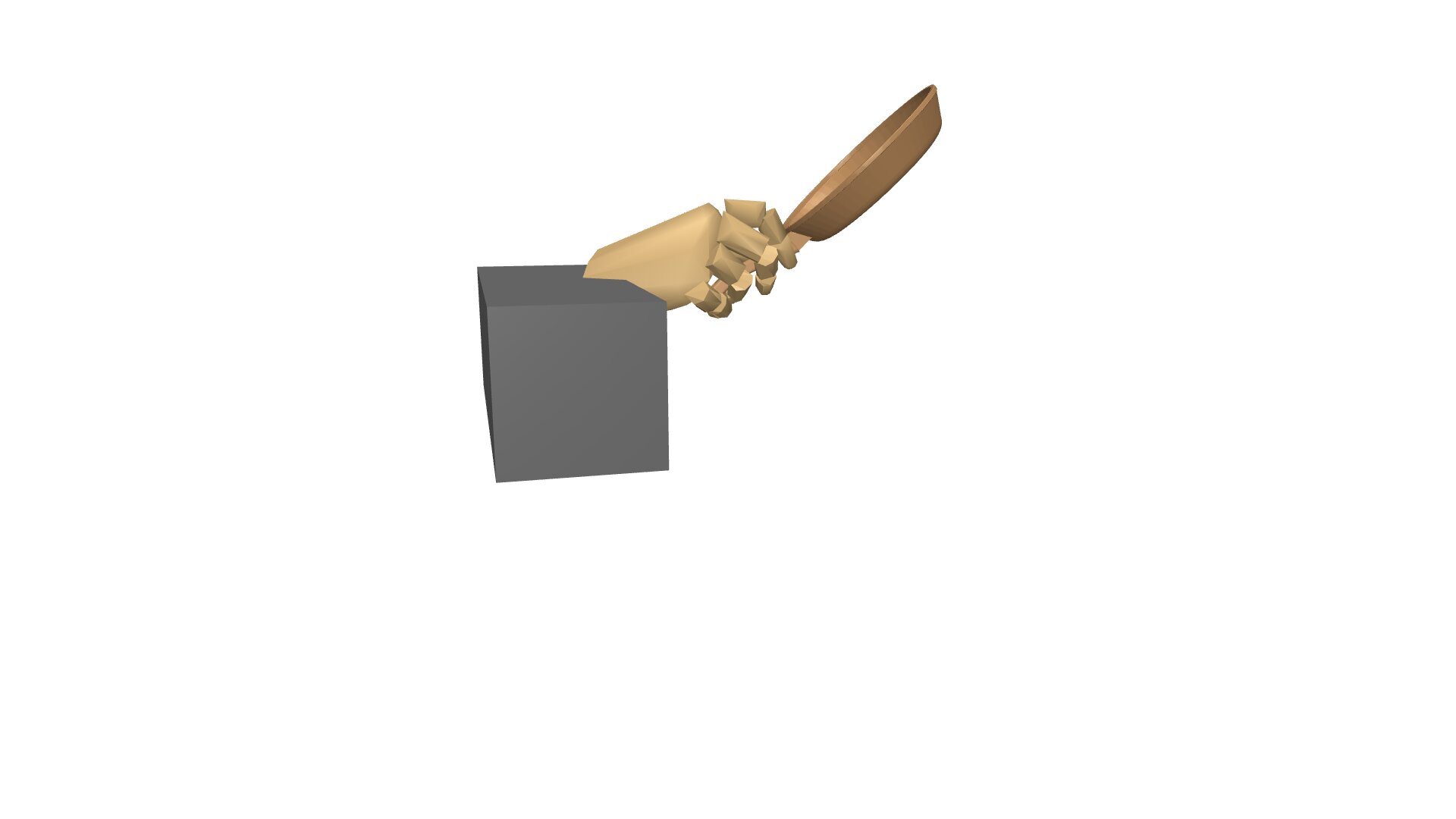}
    }\\
    \vspace{3pt}
    \subfloat[with $\textsc{loss}_\textsc{barrier}$]{
        \includegraphics[trim={24cm 14cm 20cm 4cm}, clip, width=0.92\linewidth]
        {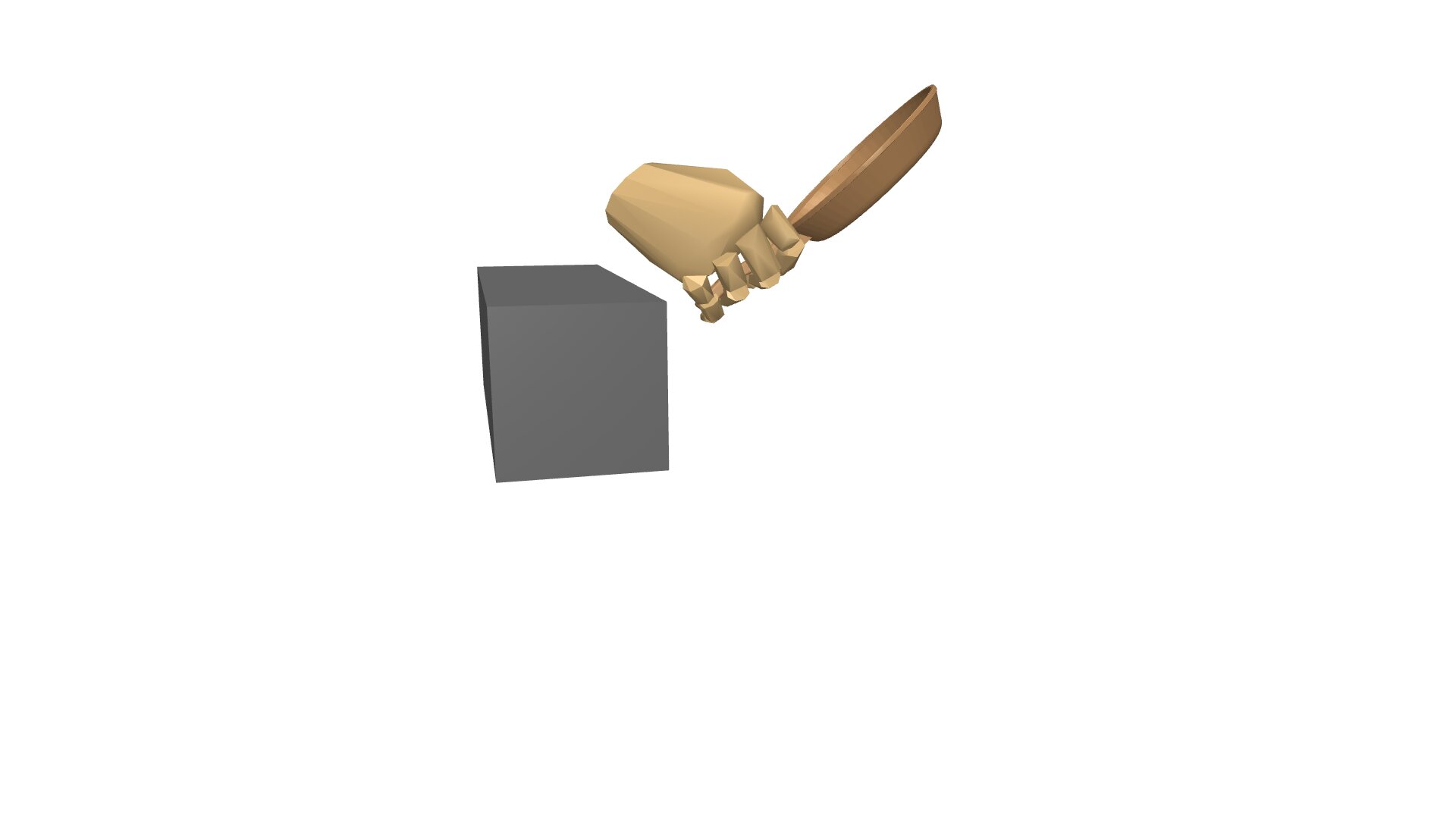}
    }
    \vspace{-10pt}
\end{wrapfigure}

\noindent where $m$ is a margin distance and $n$ controls the penalty steepness. We set $n=2.0$ and $m=0.05$\,m for all experiments. This formulation encourages a minimum clearance of $m$ from all barriers and increases sharply when the hand penetrates a barrier. We assume barrier meshes are geometric primitives, with complex meshes approximated by their collision geometries for efficient SDF evaluation. The inset figures show an example of a timestep optimized with and without $\Gamma_{B}$. We solve Eq.~\ref{eq:opt} using the PyTorch Optimization library~\cite{paszke2019pytorch}.

Finally, we perform a light smoothing with a Savitzky-Golay filter~\cite{savitzky1964smoothing} for the hand's translation and a moving-window quaternion averaging for the joint and wrist rotations. This step reduces jittering across adjacent timesteps without inducing substantial drift from the IK solution.

\section{RESULTS}

We generated 12 warping configurations of varying difficulty from 4 baseline demonstrations, which were taken from the GRAB dataset~\cite{taheri2020grab}. Contact correspondences were extracted and cleaned using \cite{lakshmipathy2025kinematic}. These configurations were made to resemble common actions in a kitchen. For instance, given a demonstration of pouring a teapot, we warp the motion to pour tea into two separate cups. The background kitchen scene is taken from the RoboCasa dataset \cite{nasiriany2024robocasa}.

For each configuration, we first warped the object trajectory to satisfy the specified waypoint and barrier constraints. We then optimized the hand trajectory at three optimization thresholds (10\,mm, 5\,mm, 3.5\,mm). At each timestep, we run Adam with a learning rate of 0.01 until the mean contact distance fell below each threshold. Each timestep was initialized using the solution from the previous timestep. All experiments were run on an Apple M2 SoC with 16 GB of memory. All runs are performed exclusively on the CPU.

\subsection{Qualitative evaluation}

Table \ref{table:2} summarizes all warping configurations used. Animations of warping results are available in the supplementary video, which show that our method successfully adapts demonstrations across a variety of constraint configurations.

\begin{table}[h!]
\centering
\caption{Evaluation trials and input specifications}
\label{table:2}
\resizebox{\columnwidth}{!} {
\begin{tabular}{l c c c c}
\hline
Trajectory & \# of waypoints & \# of barriers & $p_{\text{start}}'$ provided & $p_{\text{end}}'$ provided \\
\hline

cube\_messy & 0 & 7 & \cmark & \cmark \\
cube\_messy2 & 4 & 8 & \xmark & \cmark \\
fryingpan\_change\_tops & 1 & 0 & \cmark & \cmark \\
fryingpan\_island & 3 & 0 & \cmark & \cmark \\
fryingpan\_onto\_shelf & 4 & 3 & \cmark & \cmark \\
mug\_pass\_wall & 2 & 1 & \xmark & \xmark \\
mug\_pass\_wall\_short & 3 & 1 & \xmark & \cmark \\
mug\_pass\_in\_sink & 0 & 0 & \xmark & \cmark \\
teapot\_pour\_cup\_long & 6 & 0 & \xmark & \cmark\\
teapot\_pour\_cups & 6 & 0 & \xmark & \xmark \\
teapot\_pour\_cups\_wall & 6 & 1 & \cmark & \cmark \\
teapot\_pour\_empty & 2 & 0 & \xmark & \cmark\\
\hline
\end{tabular}}
\end{table}

\vspace{-0.2cm}

\subsection{Quantitative evaluation}

We consider two distance metrics: mean distance between the wrist position and the object's center ($D_{WO}$), and mean distance between corresponding hand and object contacts ($D_C$). The first metric measures how well our method retains the demonstration in object space, and the second metric measures the grasping accuracy after warping. Lower values indicate better performance in all tables below.

We also measure the path length ratio (PLR) by dividing the length of the warped path by the original, as well as a timewarp discrepancy metric computed as
$D_{\text{timewarp}} = \sqrt{\frac{1}{N} \sum_{i=1}^{N} \left( t_i - \frac{i-1}{N-1} \right)^2}$. The latter metric quantifies the deviation of the timewarp from uniform spacing, where a value of zero indicating no temporal distortion and larger values reflecting greater non-uniform timing.

Table \ref{table:1} reports $D_{WO}$ across all timesteps. The warped trajectories closely preserve the original hand-object spatial relationship, with $D_{WO}$ differing from the initial by just 5.5\,mm at the 10\,mm threshold, 2.2\,mm at the 5\,mm threshold, and just 1\,mm at the 3.5\,mm threshold. This demonstrates that even under substantial spatio-temporal deformations, the hand's relative positioning is maintained.

\begin{table}[h]
\centering
\vspace{0.13cm}
\caption{$D_{WO}$ across different optimization thresholds. Values are in millimeters (mm).}
\label{table:1}
\begin{tabular}{p{2.4cm} c c c c}
\hline
Trajectory & Init Mean Dist & 10\,mm & 5\,mm & 3.5\,mm \\
\hline

cube\_messy & 150.7 & 148.5 & 150.7 & 150.8 \\
cube\_messy2 & 150.7 & 149.0 & 150.0 & 149.7 \\
fryingpan\_change\_tops & 164.5 & 165.7 & 164.2 & 165.7 \\
fryingpan\_island & 164.5 & 156.6 & 159.8 & 162.6 \\
fryingpan\_onto\_shelf & 164.5 & 165.0 & 166.1 & 166.4 \\
mug\_pass\_wall & 145.4 & 143.3 & 145.3 & 145.5 \\
mug\_pass\_wall\_short & 145.4 & 146.7 & 147.1 & 147.1 \\
mug\_pass\_in\_sink & 145.4 & 147.2 & 147.2 & 147.2 \\
teapot\_pour\_cup\_long & 201.8 & 200.1 & 202.7 & 202.6 \\
teapot\_pour\_cups & 201.8 & 186.9 & 195.4 & 201.0 \\
teapot\_pour\_cups\_wall & 201.8 & 196.2 & 200.1 & 201.1 \\
teapot\_pour\_empty & 201.8 & 176.6 & 195.7 & 201.4 \\
\hline
\textbf{Average $|\Delta|$} & -- & \textbf{5.5} & \textbf{2.2} & \textbf{1.0} \\
\hline
\end{tabular}
\end{table}

Table \ref{table:3} reports $D_C$, the mean per-timestep contact distance averaged over all timesteps. At the 3.5\,mm threshold, this metric deviates from the initial value by 1.4\,mm on average, and is 1.0\,mm lower than the initial value on average. This suggests that our method preserves manipulation quality and can even achieve contact distances tighter than the original demonstration. Furthermore, post-hoc Savitzky–Golay smoothing of the hand shifts $D_C$ by only $-0.62\;$mm on average, indicating that this filter does not meaningfully erode the fidelity of per-frame IK solutions.

\begin{table}[h]
\centering
\scriptsize
\caption{$D_C$ across varied optimization thresholds in mm.}
\label{table:3}
\resizebox{\columnwidth}{!}{%
\begin{tabular}{lccccc}
\hline
Trajectory & Init Cont. & 10\,mm & 5\,mm & 3.5\,mm & Drift \\
\hline
cube\_messy              & 4.6 & 9.2 & 4.9 & 4.3 & -0.24 \\
cube\_messy2             & 4.6 & 9.4 & 6.0 & 5.5 & 0.12 \\
fryingpan\_change\_tops  & 5.5 & 7.9 & 8.5 & 5.7 & -0.37 \\
fryingpan\_island        & 5.5 & 9.2 & 6.9 & 6.1 & -0.70 \\
fryingpan\_onto\_shelf   & 5.5 & 10.0 & 6.9 & 6.1 & -0.29 \\
mug\_pass\_wall          & 8.5 & 10.6 & 6.6 & 6.1 & -0.79 \\
mug\_pass\_wall\_short   & 8.5 & 8.7 & 6.5 & 5.8 & -0.72 \\
mug\_pass\_in\_sink      & 8.5 & 10.2 & 6.0 & 5.4 & -0.88 \\
teapot\_pour\_cup\_long  & 6.4 & 7.7 & 6.3 & 4.9 & -0.74 \\
teapot\_pour\_cups       & 6.4 & 9.7 & 5.7 & 5.0 & -0.95 \\
teapot\_pour\_cups\_wall & 6.4 & 7.4 & 6.3 & 5.0 & -0.95 \\
teapot\_pour\_empty      & 6.4 & 9.3 & 6.1 & 5.0 & -0.91 \\
\hline
\textbf{Average $|\Delta|$} & -- & \textbf{2.7} & \textbf{1.3} & \textbf{1.4} & -- \\
\textbf{Average $\Delta$}   & -- & \textbf{2.7} & \textbf{0.0} & \textbf{-1.0} & -- \\
\textbf{Average drift}      & -- & -- & -- & -- & \textbf{-0.62} \\
\hline
\end{tabular}%
}
\end{table}

Table \ref{table:4} lists the time taken to run the entire pipeline. On a CPU-only architecture, the most difficult threshold (3.5\,mm) takes about 4.6 minutes to complete on average. Table \ref{table:5} reports PLR and timewarp discrepancy for all trajectories.

\begin{table}[h]
\centering
\caption{Runtime across different optimization thresholds. Non-timestep values are in seconds.}
\label{table:4}
\resizebox{\columnwidth}{!} {
\begin{tabular}{l c c c c}
\hline
Trajectory & \# of Timesteps & 10\,mm & 5\,mm & 3.5\,mm \\
\hline

cube\_messy & 673 & 74.63 & 196.93 & 464.27 \\
cube\_messy2 & 673 & 96.64 & 226.91 & 481.55 \\
fryingpan\_change\_tops & 648 & 28.39 & 80.81 & 206.03 \\
fryingpan\_island & 648 & 37.05 & 118.83 & 214.40 \\
fryingpan\_onto\_shelf & 648 & 35.59 & 110.07 & 212.03 \\
mug\_pass\_wall & 544 & 61.79 & 115.33 & 174.64 \\
mug\_pass\_wall\_short & 544 & 33.64 & 87.65 & 163.01 \\
mug\_pass\_in\_sink & 544& 42.13 & 94.98 & 148.82 \\
teapot\_pour\_cup\_long & 740 & 33.61 & 112.64  & 264.46 \\
teapot\_pour\_cups & 740 & 56.40 & 156.21 & 287.05 \\
teapot\_pour\_cups\_wall & 740 & 28.60 & 113.55 & 255.16 \\
teapot\_pour\_empty & 740 & 60.62 & 161.02 & 296.12 \\

\hline
\textbf{Average} & -- & \textbf{54.80} & \textbf{140.12} & \textbf{276.10} \\
\hline
\end{tabular}}
\end{table}

\begin{table}[h!]
\centering
\caption{PLRs and timewarp discrepencies across different optimization thresholds.}
\label{table:5}
\resizebox{\columnwidth}{!} {
\begin{tabular}{l c c c c c}
\hline
& \multicolumn{3}{c}{PLR Hand} & & \\
Trajectory & 10\,mm & 5\,mm & 3.5\,mm & PLR Object & Timewarp \\
\hline

cube\_messy & 0.999 & 1.002 & 0.989 & 1.005 & 0.075 \\
cube\_messy2 & 1.358 & 1.378 & 1.389 & 2.024 & 0.085 \\
fryingpan\_change\_tops & 1.019 & 0.945 & 0.933 & 0.897 & 0.276 \\
fryingpan\_island & 1.519 & 1.522 & 1.527 & 2.165 & 0.072 \\
fryingpan\_onto\_shelf & 1.361 & 1.314 & 1.321 & 1.733 & 0.075 \\
teapot\_pour\_empty & 1.652 & 1.649 & 1.670 & 3.688 & 0.073 \\
mug\_pass\_wall & 1.161 & 1.153 & 1.151 & 1.148 & 0.096 \\
mug\_pass\_wall\_short & 1.425 & 1.433  &  1.429 & 1.581 & 0.094  \\
mug\_pass\_in\_sink & 1.084 & 1.078 & 1.077 & 1.121 & 0.076 \\
teapot\_pour\_cup\_long &  1.590 & 1.560 & 1.574 & 2.83 &  0.139 \\
teapot\_pour\_cups & 1.059 & 1.111 & 1.155 & 1.405 & 0.117 \\
teapot\_pour\_cups\_wall &  1.542 &1.565 & 1.561 & 2.172 &  0.117 \\

\hline
\end{tabular}}
\end{table}

\begin{figure}[h]%
    \centering
    {\includegraphics[width=0.97\linewidth]{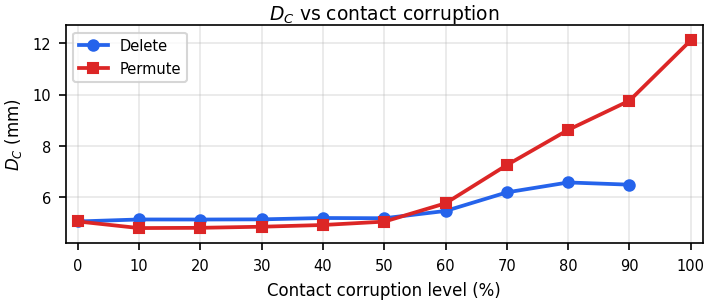}}
    \caption{Contact correspondence robustness test on the teapot\_pour\_cups\_wall scene. Both curves maintain correspondence quality up until 50\% ($D_c$ stays near 5.0 mm).}%
    \label{fig:robustness}%
\end{figure}

Figure \ref{fig:robustness} tests robustness on the teapot\_pour\_cups\_wall scene by corrupting demonstration contacts in two ways: deleting contacts and randomly permuting object-vertex labels among contact pairs. Both remain largely stable through 50\%; beyond that, $D_C$ rises as corruption increases and retargeting quality degrades, with the effect more pronounced at higher corruption levels. Permuting is more severe post-50\% because it mis-specifies targets, pulling the hand toward incorrect object points; deleting only removes constraints, and since each frame is initialized from the previous pose, the trajectory is more resistant to degradation.

\vspace{-0.25cm}

\subsection{Generalization}

Although our experiments consider a dexterous human hand and a human motion dataset, our method is not specific to any end-effector. We can easily use our approach on parallel-jaw grippers and multi-fingered robot platforms.

\subsubsection{Franka arm with parallel grippers}
Given a task demonstration, we can trivially compute two contact points (one for each jaw where it intersects with the object).

\begin{figure}[h]%
    \centering
    \subfloat[\centering input trajectory]{{\includegraphics[trim={17cm 1cm 25cm 8cm}, clip, width=3.84cm]{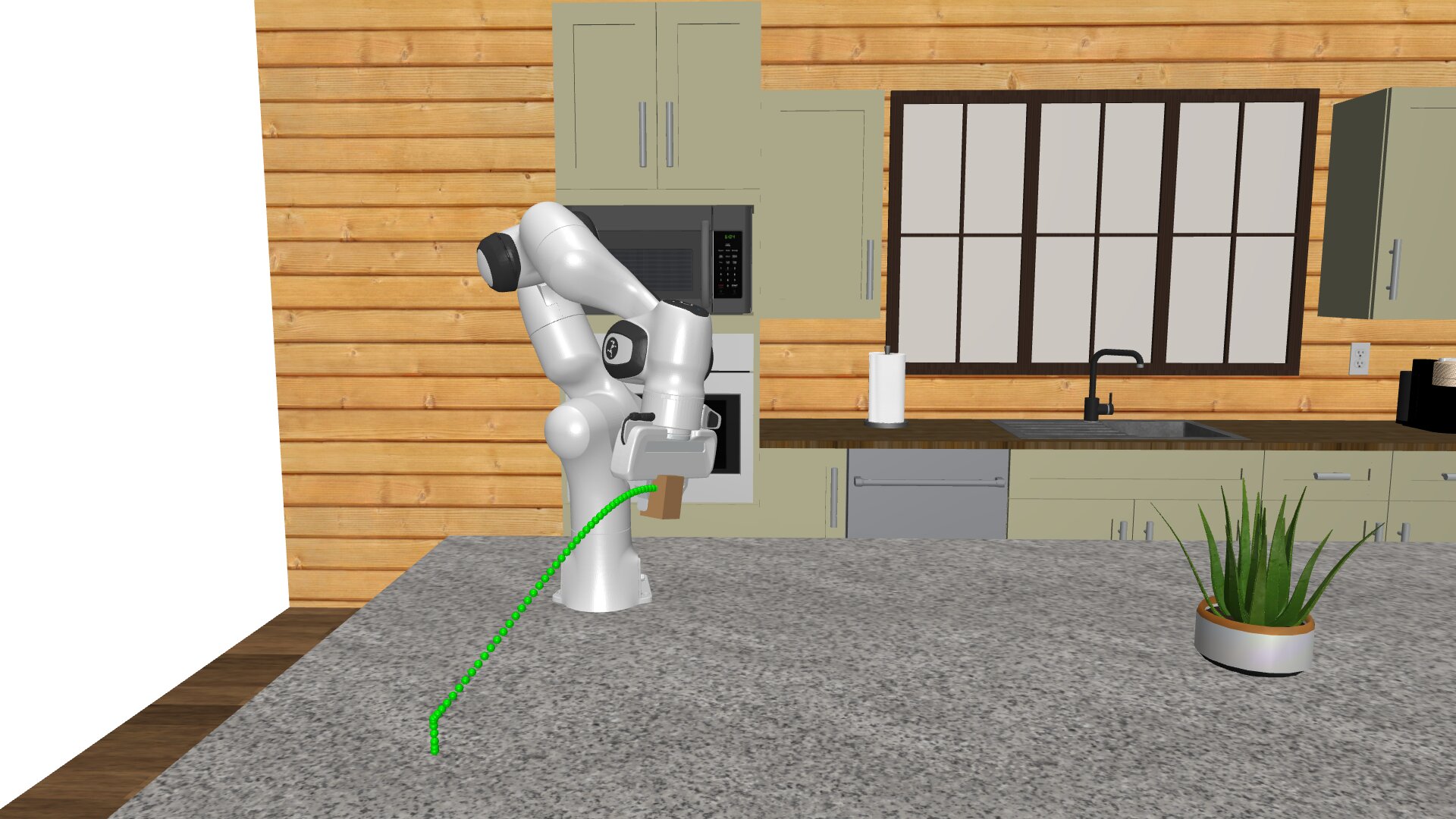} }}%
    \qquad
    \subfloat[\centering warped trajectory]{{\includegraphics[trim={17cm 1cm 25cm 8cm}, clip, width=3.84cm]{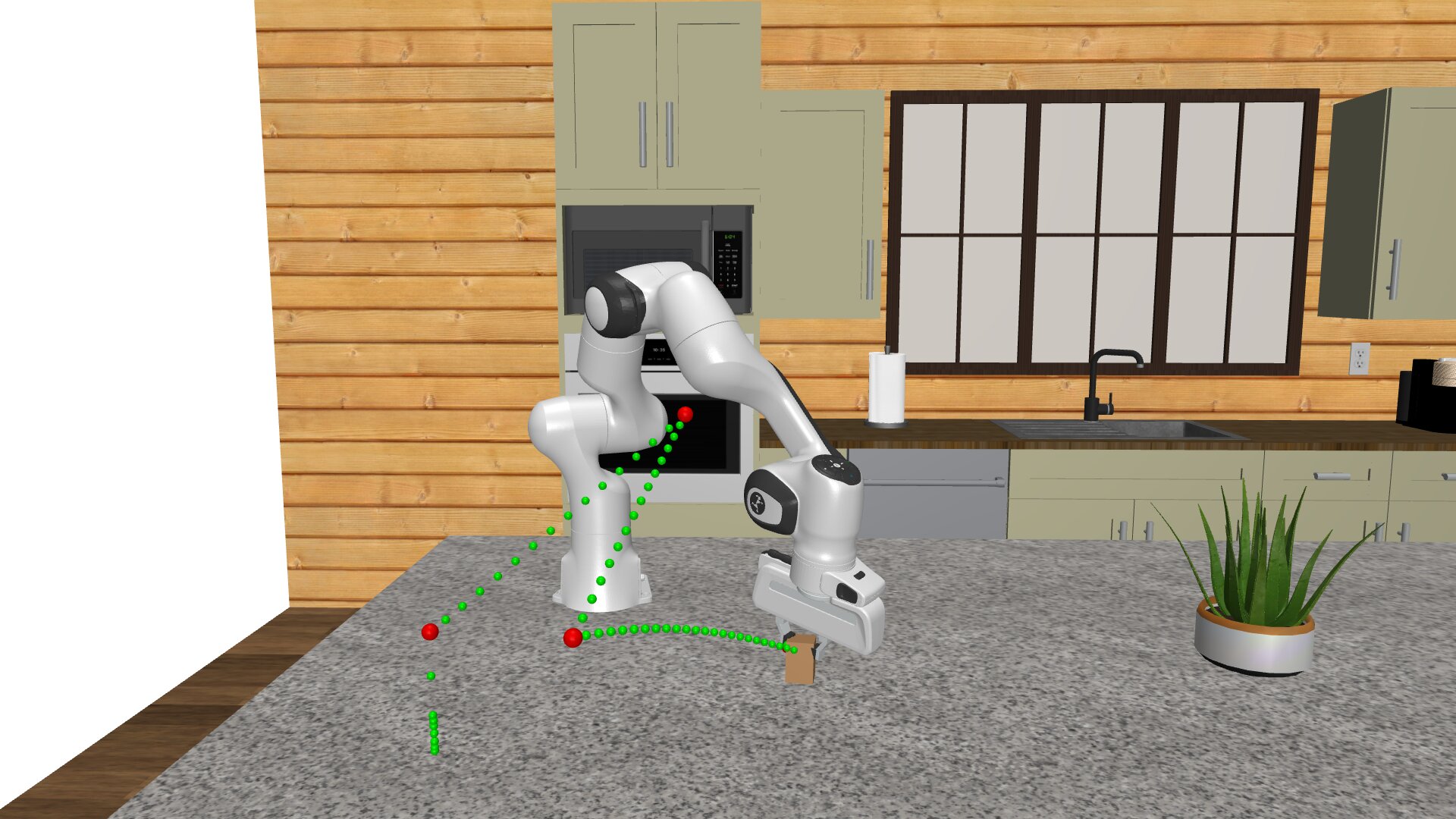} }}%
    \caption{An example warping using the Franka arm. The hand must move the box through a series of waypoints.}%
    \label{fig:franka}%
\end{figure}

Figure \ref{fig:franka} shows that this is enough to successfully warp a Franka gripper demonstration. In this example, we begin with a demonstration in which the gripper lifts and moves a small box. The trajectory is warped to move the box through a series of waypoints and then place it on the table.

\subsubsection{Allegro hand}
\begin{figure}[h!]%
    \centering
    % \vspace{0.16cm}
    \subfloat[\centering input trajectory]{{\includegraphics[trim={20cm 1cm 22cm 0cm}, clip, height=5.4cm]{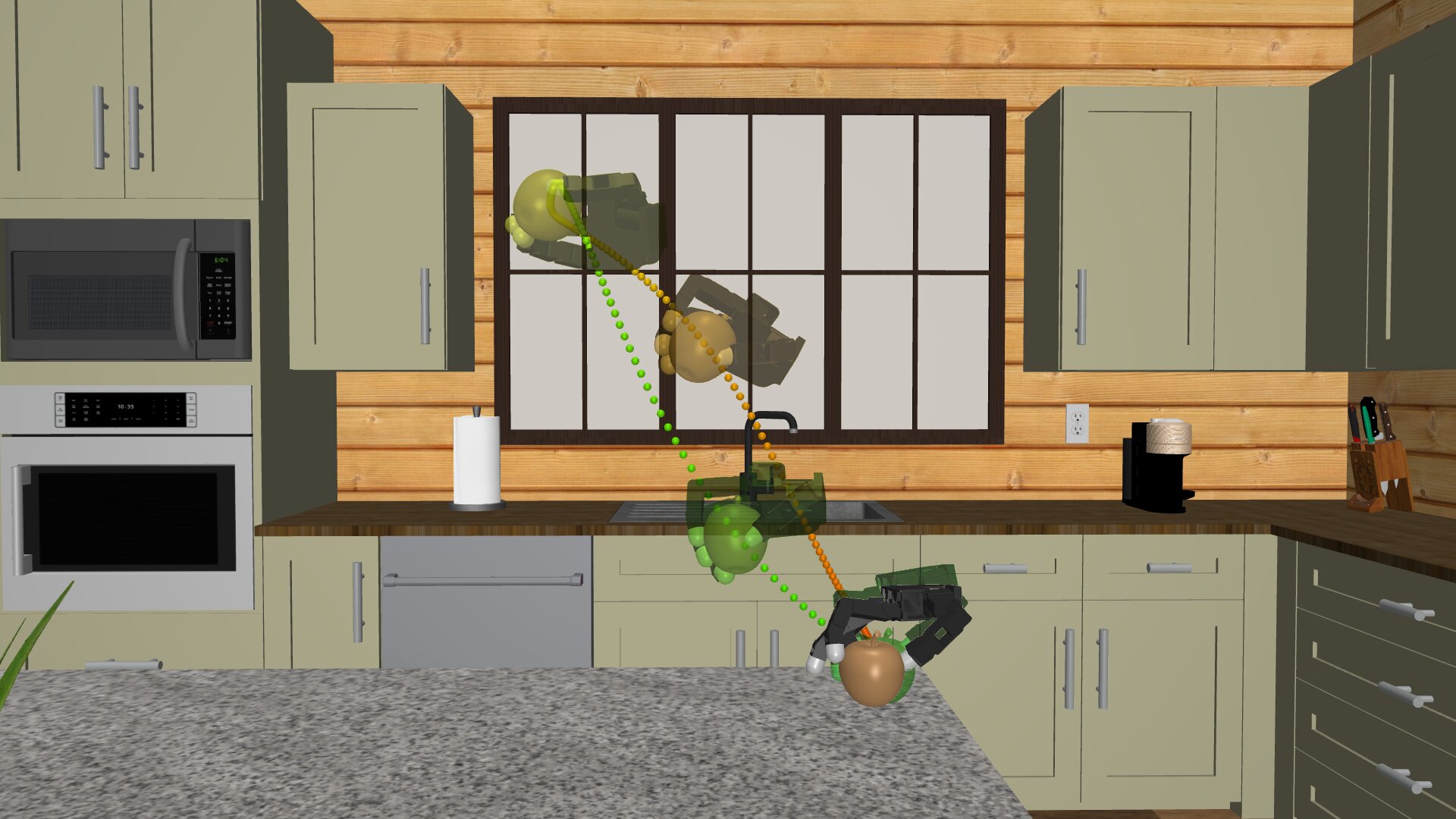} }}%
    \qquad
    \subfloat[\centering warped trajectory]{{\includegraphics[trim={22cm 1cm 19cm 0cm}, clip, height=5.4cm]{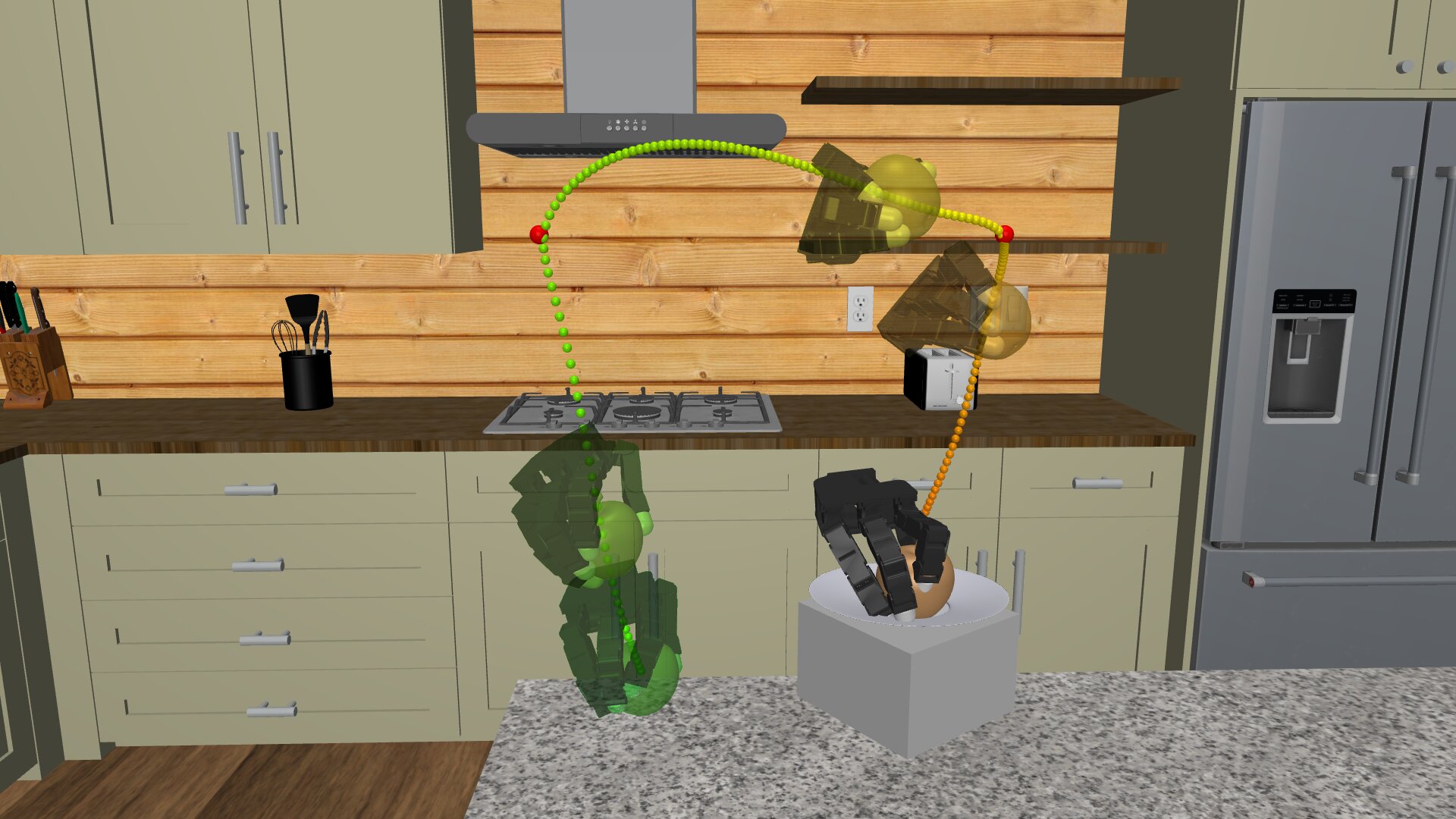} }}%
    \qquad
    \subfloat[\centering timewarp]{{\includegraphics[trim={0cm 35cm 0cm 0cm}, clip, width=8.5cm]{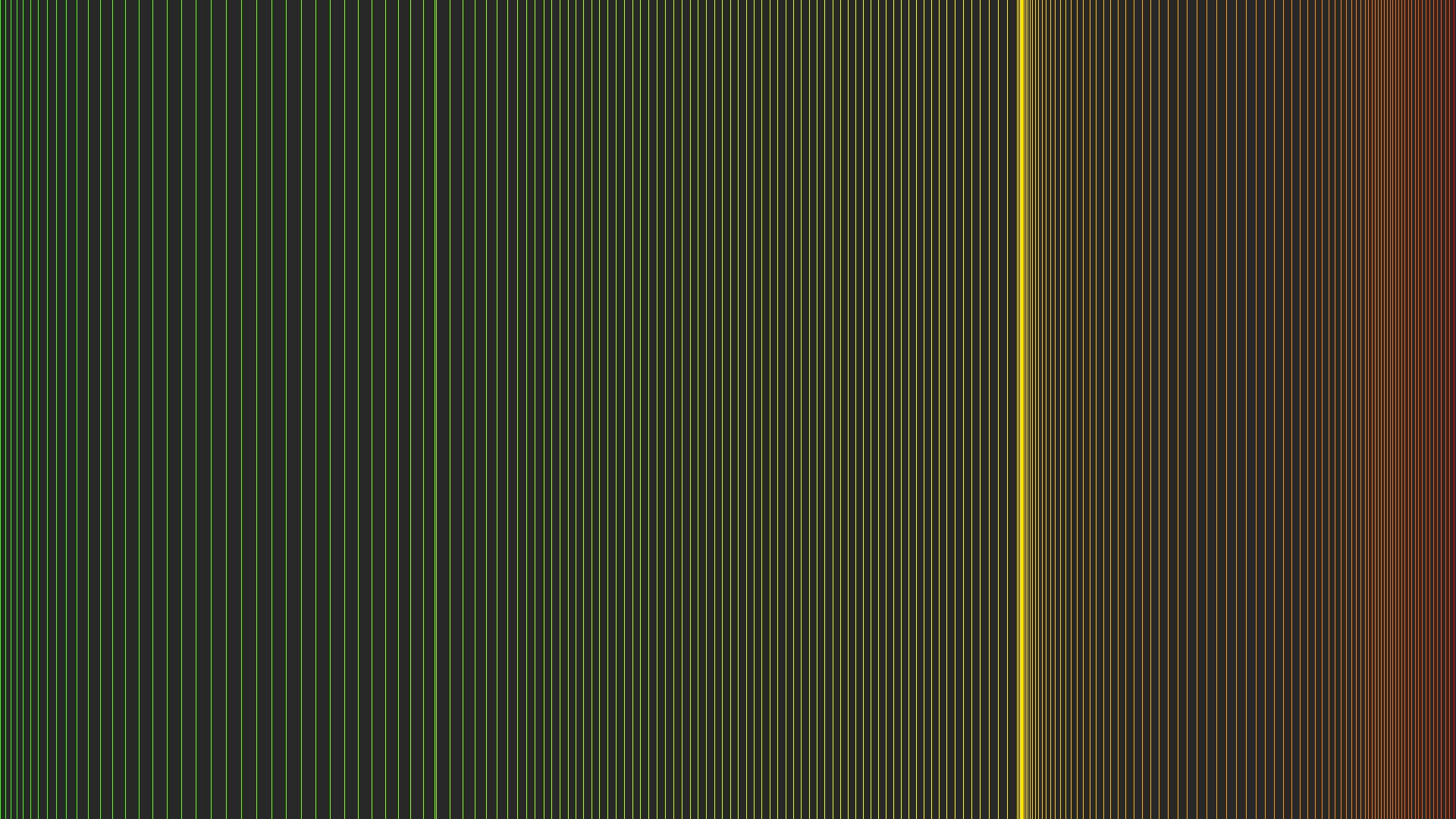} }}
    \caption{An example Allegro hand warping where (a) the input apple handoff trajectory is (b) altered to pass through waypoints (red) and terminate on an elevated plate. (c) The timewarp visualization illustrates where frames of the original trajectory fall in the warped sequence, with black portions indicating segments which must be interpolated.}
    \label{fig:allegro}%
    \vspace{-0.2cm}
\end{figure}

Figure \ref{fig:allegro} shows that the method generalizes to other dexterous hands such as Allegro. The initial trajectory simply picks up the apple, and our method warps it to move the apple through waypoints and place it on an elevated plate. The resulting timewarp is visualized in Figure \ref{fig:allegro}(c), showing the non-uniform temporal redistribution induced by the warping. Despite significant warping, $D_{WO}$ changes only marginally (0.133\;m $\rightarrow$ 0.137\;m), suggesting that our method preserves the hand–object spatial relationship.

\subsection{Baseline comparison}
\label{sec:baseline}

Previous demonstration augmentation methods rigidly adapt hand trajectories using a constant $SE(3)$ transformation applied to all timesteps~\cite{mandlekar2023mimicgen, jiang2025dexmimicgen}. To construct a comparable baseline for our setting, we extend this approach by computing rigid-body transforms at the input timesteps that the timewarp maps directly to output timesteps.

However, because we only know the true hand-object relationship at input timesteps, the only ``correct" hand configurations are in $\tau$'s codomain. We therefore must interpolate poses at timesteps where no true configuration exists. To do so, we treat these known configurations as control points, fit a B-spline parameterized by the timewarp, and sample it at equal increments to produce the baseline hand trajectory.

This approach introduces a fundamental problem. While the hand-object relationship is preserved at the control point timesteps, the interpolated hand positions between them are independent of the object's position. This issue is especially exacerbated when the timewarp is non-uniform or when the spatial deformation is large. In contrast, our method explicitly optimizes the hand pose at every output timestep to minimize contact point distances, ensuring the hand-object relationship is preserved throughout the entire trajectory rather than just at samples from the original trajectory.

\begin{figure*}[h]
    \centering
    \vspace{0.16cm}
    \includegraphics[width=1.6\columnwidth]{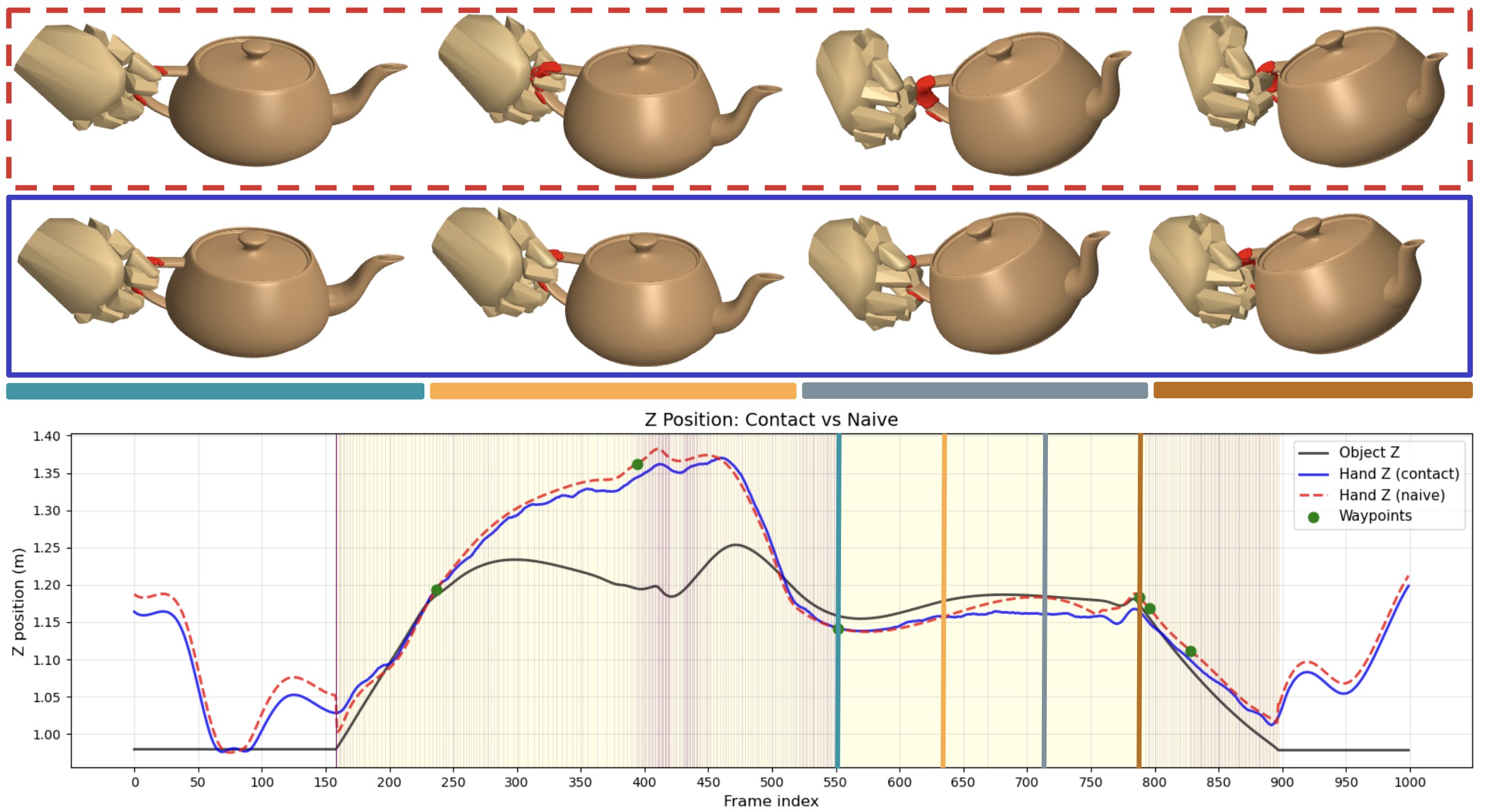}
    \caption{Comparison of the baseline and our hand warping methods. The yellow region marks the in-contact segment. Light purple lines show the timewarp mapping from input to output timesteps. Four timesteps are visualized for both warping methods (cyan, orange, gray, brown). The baseline method degrades at the orange and gray timesteps, where ground-truth transformation is not available. Our contact-based method continues to maintain a proper grasp across all timesteps.}
    \label{fig:naive}
    \vspace{-9pt}
\end{figure*}

Figure \ref{fig:naive} shows an example failure case of the naive version. The graph depicts the z-position of the warped object trajectory, as well as the naive and contact-based hand warped trajectories. The yellow region marks the contact region, and the purple lines represent the timesteps mapped from the input trajectory to the output trajectory by $\tau$.

Between timesteps 529 and 750, there are few purple lines, indicating significant temporal stretching from the input trajectory in this region. Here, the naive method must interpolate the hand trajectory, and regardless of the interpolation scheme used (cubic in this case), the resulting hand motion diverges from the object's path.

At timesteps 528 and 751 where we have the ground truth hand-object relationship, the naive method's hand maintains a proper grasp of the object. However, at interpolated points, the method fails to extrapolate the correct hand transformation, leading to intersections (orange) or failure to maintain contact (gray). Our contact-based method directly combats interpolation errors, leading to a more accurate warping.

\section{DISCUSSION}

% We briefly discuss advantages, limitations, and failure cases of our method.

\subsection{Advantages and Disadvantages}

Our contact-based method is especially beneficial when the warping configuration introduces substantial spatio-temporal deformation (e.g., far-away waypoints or highly non-linear timewarp). Our approach is also particularly well-suited to interactions where precise hand-object contact is critical, such as grasping, pouring, or tool use. Additionally, many existing methods do not have built-in mechanisms for resolving hand-barrier collisions. Our method naturally handles this through the barrier loss term during per-timestep optimization. This implies our approach is especially valuable in cluttered scenes with multiple obstacles. However, we do note that mild warpings, especially those without temporal shifts or barriers, may be serviced by faster techniques~\cite{mandlekar2023mimicgen}.

\subsection{Limitations}

Our method currently has a number of limitations. First, we assume that all objects are rigid bodies. We do not support articulated or deformable objects, which would require significant computation to track the motion of individual components during both object and hand warping.

We assume that all barriers are convex. Concave barriers may therefore be too conservatively approximated.

Additionally, our method is evaluated purely in simulation and in isolation rather than as a component in a larger learning pipeline. Although our goal in this work is to focus on the core warp operation, which we view as an upstream processing problem, it is not clear how effective these warps are in sim2real deployment pipelines. A comprehensive investigation into large-scale domain randomization using our approach is an exciting direction for future work.

%\hl{While our method is primarily designed for offline data augmentation, if online deployment were desired, two bottlenecks arise: per-frame IK optimization and contact extraction. The former may be mitigated by relaxing the optimization threshold. A comprehensive evaluation in a closed-loop setting remains an interesting direction for future work.}

% Moreover, our evaluation is limited to a small number of warping configurations. However, we note that the selected configurations were designed to cover a diverse set of constraint combinations: varying the number of waypoints (0-6) and barriers (0-8), differing magnitudes and directions of spatial, rotational, and temporal constraints, severe non-uniform temporal retimings, varying the shape, scale, and placement of barrier meshes, and varying initial and final displacements. Together, these combinations exercise all components of the pipeline and demonstrate its robustness across a representative set of warping configurations.

%\hl{The recovered hand trajectory's quality is directly bounded by the accuracy of the input contact correspondences. When few contacts are noisy, per-frame IK optimization partially absorbs mild inaccuracies by minimizing mean contact distance. Severe contact inaccuracies are a known failure case.}

Finally, our method does not interpolate contact correspondence between timesteps, instead reusing the last available set. This restricts the types of tasks that can be effectively warped. Warping demonstrations involving rapid finger movements or fine-grained intermediate contact changes, for example, may fail as the evolving contact distribution is not captured between available reference timesteps. Formulating simple input specifications that build upon contact interpolation works~\cite{lakshmipathy2021contact} may address this limitation.

\subsection{Failure cases}

While generally effective, our method does exhibit failure cases which we discuss below.

\subsubsection{Hand-barrier collisions}
% \begin{wrapfigure}{r}{0.04\textwidth}
%     \vspace{-20pt}
%     \subfloat{{\includegraphics[trim={23cm 2cm 25cm 18cm}, clip, width=2.6cm]{fryingpan_into_dishwasher_hand_barrier_contact_frame_0330_top.jpg} }}\vspace{5pt}\\
%     \subfloat{{\includegraphics[trim={23cm 16cm 25cm 7cm}, clip, width=2.6cm]{fryingpan_into_dishwasher_hand_barrier_contact_frame_0330.jpg} }}%
%     \vspace{-10pt}
% \end{wrapfigure}
\begin{figure}[h]
    \centering
    \vspace{-10pt}
    \begin{minipage}[b]{2.6cm}
        \centering
        \includegraphics[trim={23cm 2cm 25cm 18cm}, clip, width=2.6cm]{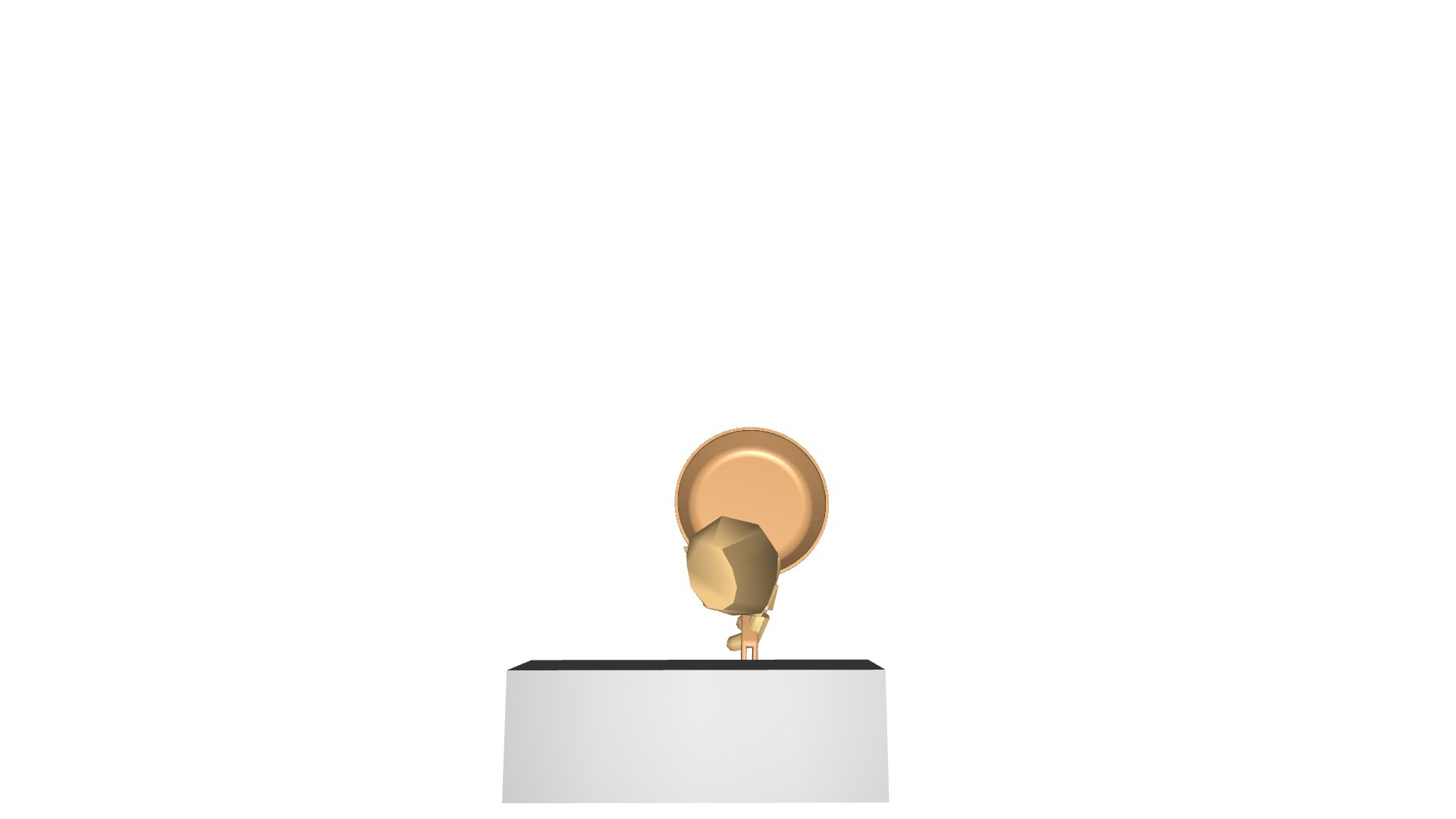}\\[2pt]
        \small (a) top view
    \end{minipage}
    \hspace{10pt}
    \begin{minipage}[b]{2.6cm}
        \centering
        \includegraphics[trim={23cm 16cm 25cm 7cm}, clip, width=2.6cm]{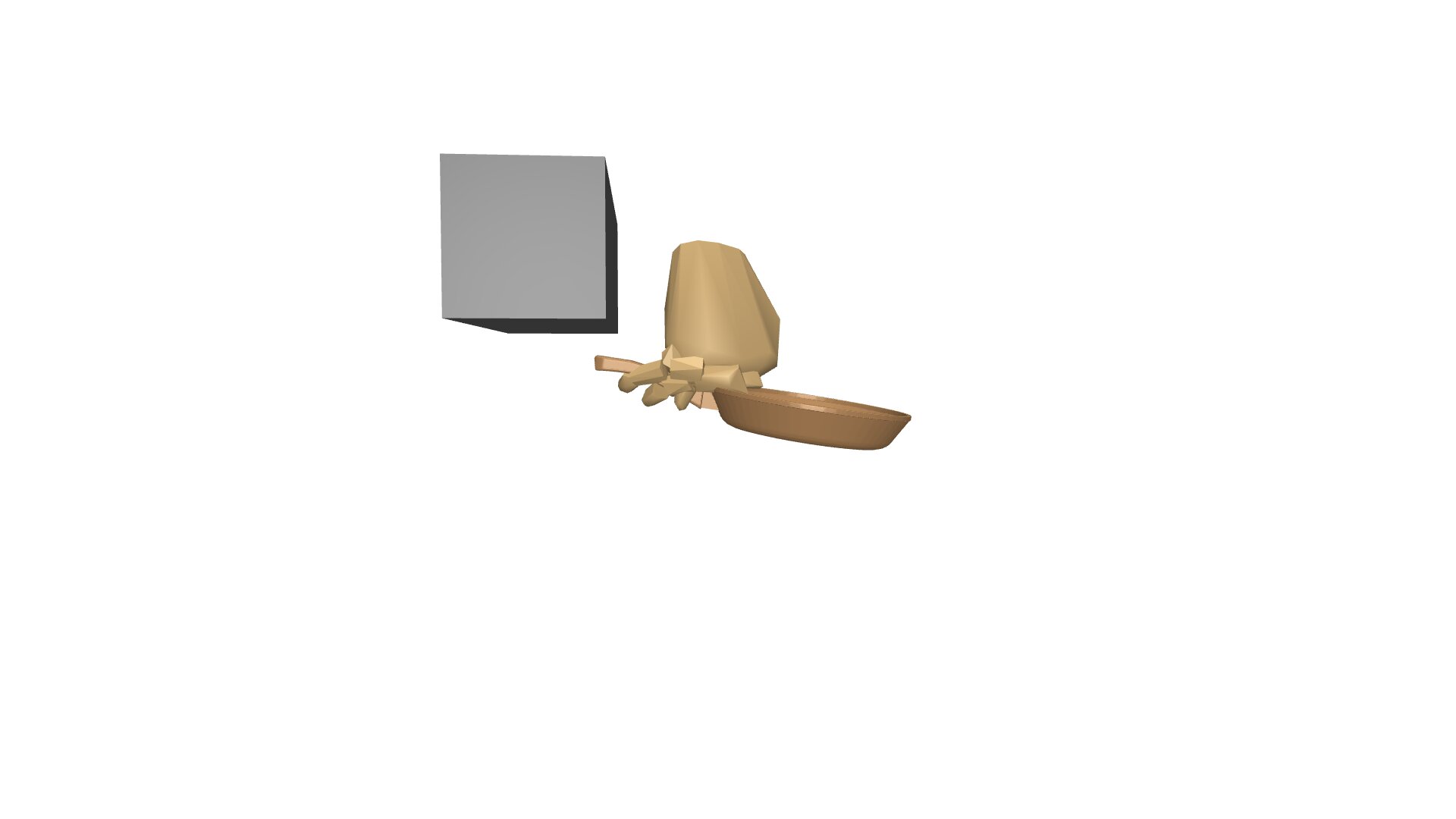}\\[2pt]
        \small (b) side view
    \end{minipage}
    \vspace{-3pt}
\end{figure}

The warped object trajectory may navigate past barriers such that the hand cannot maintain a proper grasp of the object without intersecting a barrier. The fryingpan inset shows such an instance, where the object barely clears the corner of a box, leaving insufficient space for the hand. These cases can be mitigated by increasing the object's bounding ball to provide additional clearance.

\subsubsection{Narrow gaps}

\begin{wrapfigure}{r}{2.6cm}
    \vspace{-10pt}
    % \subfloat{{\includegraphics[trim={23cm 5.5cm 19.5cm 12cm}, clip, width=2.6cm]{mug_pass_narrow_waypoints_only_frame_0790.jpg} }}\\
    \subfloat{{\includegraphics[trim={23cm 5.5cm 19.5cm 2.5cm}, clip, width=2.3cm]{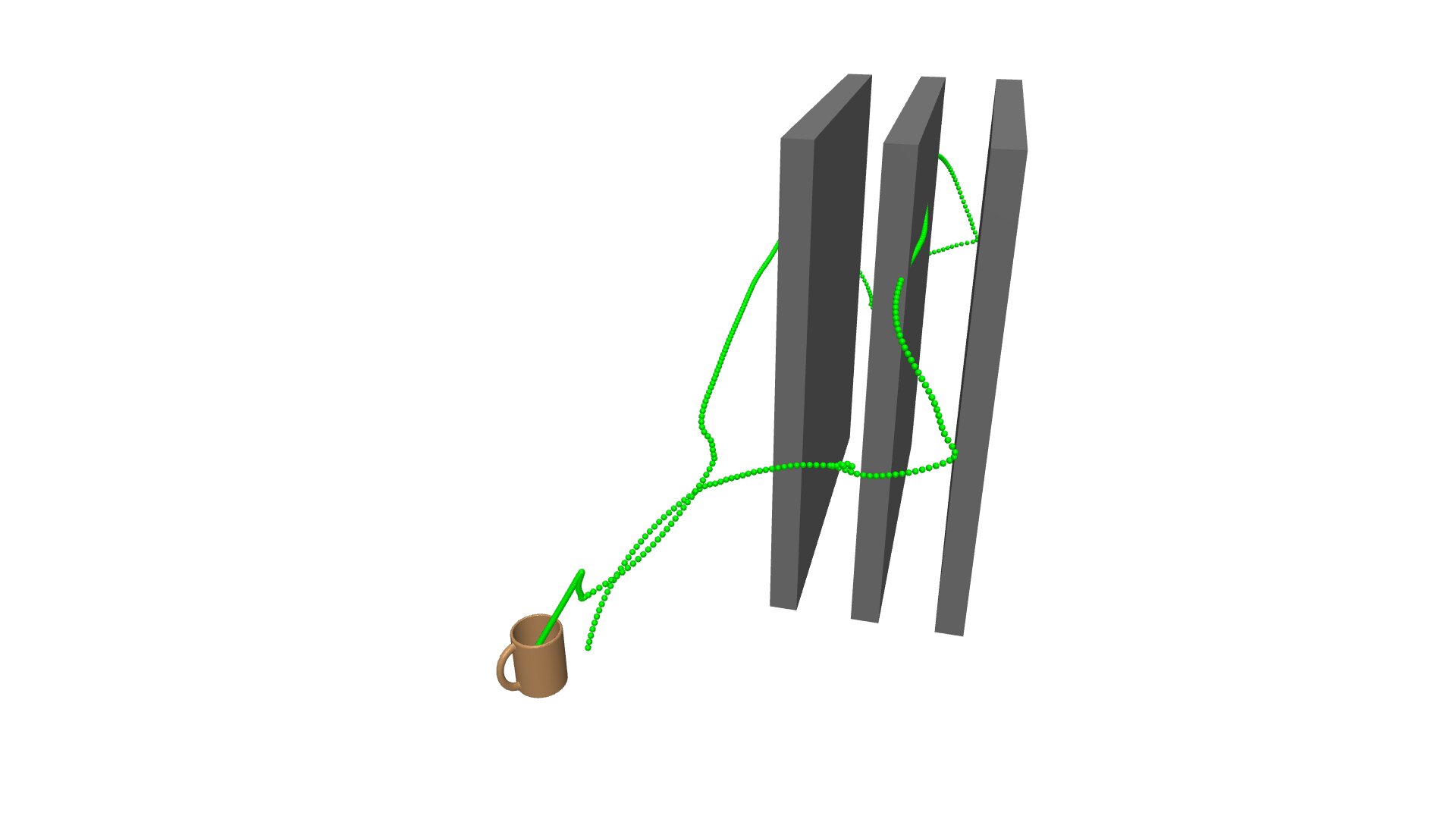} }}
    \vspace{-10pt}
\end{wrapfigure}

Barrier-object collisions are resolved iteratively, so pushing trajectory points out of one barrier may displace them into an adjacent one. The right inset shows such a case, where barriers form narrow gaps, causing the barrier resolution to oscillate trajectory points without converging to a valid trajectory.

\section{CONCLUSION}

We have presented a straightforward but effective method for nonlinear spatio-temporal warping of contact-rich dexterous manipulations. Our framework effectively reconciles spatio-temporal waypoints and environmental barriers with varying start-to-end configurations. Using hand-object contact distributions, we successfully decomposed the task into an object-centric trajectory pipeline followed by contact-preserving hand warping via per-timestep IK optimization. Our evaluation shows that the warped trajectories preserve the hand-object relationship to within 1\,mm on average, while being computationally tractable even on a CPU, and that our method generalizes across different end-effectors, including a parallel-jaw gripper and the Allegro hand. Our method is well poised to serve as a modular, practical upstream component in demonstration augmentation pipelines, enabling efficient generation of spatially and temporally diverse training data. We hope that the simplicity of our approach, together with a public code release, will encourage its adoption within data augmentation pipelines.

\section*{ACKNOWLEDGMENTS}
This research was partially supported by the National Institute of Biomedical Imaging and Bioengineering of the National Institutes of Health under Award Number R01EB036842, and by the RAI Institute.

\bibliographystyle{IEEEtran}
\bibliography{references}

@article{lakshmipathy2025kinematic,
  title={Kinematic motion retargeting for contact-rich anthropomorphic manipulations},
  author={Lakshmipathy, Arjun S and Hodgins, Jessica K and Pollard, Nancy S},
  journal={ACM Transactions on Graphics},
  volume={44},
  number={2},
  pages={1--20},
  year={2025},
  publisher={ACM New York, NY}
}

@article{chen2023genaug,
  title={Genaug: Retargeting behaviors to unseen situations via generative augmentation},
  author={Chen, Z. and Kiami, S. and Gupta, A. and Kumar, V.},
  journal={arXiv preprint arXiv:2302.06671},
  year={2023}
}

@inproceedings{bharadhwaj2024roboagent,
  title={Roboagent: Generalization and efficiency in robot manipulation via semantic augmentations and action chunking},
  author={Bharadhwaj, Homanga and Vakil, Jay and Sharma, Mohit and Gupta, Abhinav and Tulsiani, Shubham and Kumar, Vikash},
  booktitle={IEEE International Conference on Robotics and Automation},
  pages={4788--4795},
  year={2024},
  organization={IEEE}
}

@article{oh2025self,
  title={Self-Augmented Robot Trajectory: Efficient Imitation Learning via Safe Self-augmentation with Demonstrator-annotated Precision},
  author={Oh, Hanbit and Murooka, Masaki and Motoda, Tomohiro and Nakajo, Ryoichi and Domae, Yukiyasu},
  journal={arXiv preprint arXiv:2509.09893},
  year={2025}
}

@article{xue2025demogen,
  title={Demogen: Synthetic demonstration generation for data-efficient visuomotor policy learning},
  author={Xue, Zhengrong and Deng, Shuying and Chen, Zhenyang and Wang, Yixuan and Yuan, Zhecheng and Xu, Huazhe},
  journal={arXiv preprint arXiv:2502.16932},
  year={2025}
}

@article{mandlekar2023mimicgen,
  title={Mimicgen: A data generation system for scalable robot learning using human demonstrations},
  author={Mandlekar, Ajay and Nasiriany, Soroush and Wen, Bowen and Akinola, Iretiayo and Narang, Yashraj and Fan, Linxi and Zhu, Yuke and Fox, Dieter},
  journal={arXiv preprint arXiv:2310.17596},
  year={2023}
}

@inproceedings{jiang2025dexmimicgen,
  title={Dexmimicgen: Automated data generation for bimanual dexterous manipulation via imitation learning},
  author={Jiang, Zhenyu and Xie, Yuqi and Lin, Kevin and Xu, Zhenjia and Wan, Weikang and Mandlekar, Ajay and Fan, Linxi Jim and Zhu, Yuke},
  booktitle={2025 IEEE International Conference on Robotics and Automation (ICRA)},
  pages={16923--16930},
  year={2025},
  organization={IEEE}
}

@inproceedings{cheng2022contact,
  title={Contact mode guided motion planning for quasidynamic dexterous manipulation in 3d},
  author={Cheng, Xianyi and Huang, Eric and Hou, Yifan and Mason, Matthew T},
  booktitle={IEEE International Conference on Robotics and Automation},
  pages={2730--2736},
  year={2022}
}

@inproceedings{kalakrishnan2011stomp,
  title={STOMP: Stochastic trajectory optimization for motion planning},
  author={Kalakrishnan, Mrinal and Chitta, Sachin and Theodorou, Evangelos and Pastor, Peter and Schaal, Stefan},
  booktitle={IEEE international conference on robotics and automation},
  pages={4569--4574},
  year={2011}
}

@inproceedings{ratliff2009chomp,
  title={CHOMP: Gradient optimization techniques for efficient motion planning},
  author={Ratliff, Nathan and Zucker, Matt and Bagnell, J Andrew and Srinivasa, Siddhartha},
  booktitle={2009 IEEE international conference on robotics and automation},
  pages={489--494},
  year={2009},
  organization={IEEE}
}

@article{jin2025physically,
  title={Physically-based Lighting Generation for Robotic Manipulation},
  author={Jin, S. and Wang, L. and Temming, B. and Pokorny, F. T.},
  journal={arXiv preprint arXiv:2508.01442},
  year={2025}
}

@inproceedings{qureshi2019motion,
  title={Motion planning networks},
  author={Qureshi, Ahmed H and Simeonov, Anthony and Bency, Mayur J and Yip, Michael C},
  booktitle={2019 International Conference on Robotics and Automation (ICRA)},
  pages={2118--2124},
  year={2019},
  organization={IEEE}
}

@article{nasiriany2024robocasa,
  title={Robocasa: Large-scale simulation of everyday tasks for generalist robots},
  author={Nasiriany, Soroush and Maddukuri, Abhiram and Zhang, Lance and Parikh, Adeet and Lo, Aaron and Joshi, Abhishek and Mandlekar, Ajay and Zhu, Yuke},
  journal={arXiv preprint arXiv:2406.02523},
  year={2024}
}

@inproceedings{fan2023arctic,
    title = {{ARCTIC}: A Dataset for Dexterous Bimanual Hand-Object Manipulation},
    author = {Fan, Z. and Taheri, O. and Tzionas, D. and Kocabas, M. and Kaufmann, M. and Black, M. J. and Hilliges, O.},
    booktitle = {IEEE/CVF Conference on Computer Vision and Pattern Recognition},
    pages={12943--12954},
    year = {2023}
}

@inproceedings{taheri2020grab,
  title={GRAB: A dataset of whole-body human grasping of objects},
  author={Taheri, Omid and Ghorbani, Nima and Black, Michael J and Tzionas, Dimitrios},
  booktitle={European conference on computer vision},
  pages={581--600},
  year={2020},
  organization={Springer}
}

@inproceedings{christen2022d,
  title={D-grasp: Physically plausible dynamic grasp synthesis for hand-object interactions},
  author={Christen, Sammy and Kocabas, Muhammed and Aksan, Emre and Hwangbo, Jemin and Song, Jie and Hilliges, Otmar},
  booktitle={Proceedings of the IEEE/CVF Conference on Computer Vision and Pattern Recognition},
  pages={20577--20586},
  year={2022}
}

@inproceedings{wu2022saga,
  title={Saga: Stochastic whole-body grasping with contact},
  author={Wu, Yan and Wang, Jiahao and Zhang, Yan and Zhang, Siwei and Hilliges, Otmar and Yu, Fisher and Tang, Siyu},
  booktitle={European Conference on Computer Vision},
  pages={257--274},
  year={2022},
  organization={Springer}
}

@article{ye2012synthesis,
  title={Synthesis of detailed hand manipulations using contact sampling},
  author={Ye, Yuting and Liu, C Karen},
  journal={ACM Transactions on Graphics (ToG)},
  volume={31},
  number={4},
  pages={1--10},
  year={2012},
  publisher={ACM New York, NY, USA}
}

@article{hazard2020automated,
  title={Automated design of robotic hands for in-hand manipulation tasks},
  author={Hazard, Christopher and Pollard, Nancy and Coros, Stelian},
  journal={International Journal of Humanoid Robotics},
  volume={17},
  number={01},
  pages={1950029},
  year={2020},
  publisher={World Scientific}
}

@inproceedings{brahmbhatt2019contactgrasp,
  title={Contactgrasp: Functional multi-finger grasp synthesis from contact},
  author={Brahmbhatt, Samarth and Handa, Ankur and Hays, James and Fox, Dieter},
  booktitle={2019 IEEE/RSJ International Conference on Intelligent Robots and Systems (IROS)},
  pages={2386--2393},
  year={2019},
  organization={IEEE}
}

@inproceedings{turpin2022grasp,
  title={Grasp’d: Differentiable contact-rich grasp synthesis for multi-fingered hands},
  author={Turpin, Dylan and Wang, Liquan and Heiden, Eric and Chen, Yun-Chun and Macklin, Miles and Tsogkas, Stavros and Dickinson, Sven and Garg, Animesh},
  booktitle={European Conference on Computer Vision},
  pages={201--221},
  year={2022},
  organization={Springer}
}

@article{le2024fast,
  title={Fast contact-implicit model predictive control},
  author={Le Cleac'h, S. and Howell, T. A. and Yang, S. and Lee, C. Y. and Zhang, J. and Bishop, A. and Schwager, M. and Manchester, Z.},
  journal={IEEE Transactions on Robotics},
  volume={40},
  pages={1617--1629},
  year={2024},
  publisher={IEEE}
}

@article{pang2023global,
    title={Global planning for contact-rich manipulation via local smoothing of quasi-dynamic contact models},
    author={Pang, T. and Suh, H. J. T. and Yang, L. and Tedrake, R.},
    journal={IEEE Transactions on robotics},
    volume={39},
    number={6},
    pages={4691--4711},
    year={2023},
    publisher={IEEE}
}

@article{suh2025dexterous,
  title={Dexterous contact-rich manipulation via the contact trust region},
  author={Suh, HJ Terry and Pang, Tao and Zhao, Tong and Tedrake, Russ},
  journal={The International Journal of Robotics Research},
  pages={02783649251398875},
  year={2025},
  publisher={SAGE Publications Sage UK: London, England}
}

@article{zhang2025motion,
  title={Motion planning for robotics: A review for sampling-based planners},
  author={Zhang, L. and Cai, K. and Sun, Z. and Bing, Z. and Wang, C. and Figueredo, L. and Haddadin, S. and Knoll, A.},
  journal={Biomimetic Intelligence and Robotics},
  volume={5},
  number={1},
  pages={100207},
  year={2025},
  publisher={Elsevier}
}

@inproceedings{carvalho2023motion,
  title={Motion planning diffusion: Learning and planning of robot motions with diffusion models},
  author={Carvalho, Joao and Le, An T and Baierl, Mark and Koert, Dorothea and Peters, Jan},
  booktitle={2023 IEEE/RSJ International Conference on Intelligent Robots and Systems (IROS)},
  pages={1916--1923},
  year={2023},
  organization={IEEE}
}

@inproceedings{seo2025presto,
  title={Presto: Fast motion planning using diffusion models based on key-configuration environment representation},
  author={Seo, Mingyo and Cho, Yoonyoung and Sung, Yoonchang and Stone, Peter and Zhu, Yuke and Kim, Beomjoon},
  booktitle={2025 IEEE International Conference on Robotics and Automation (ICRA)},
  pages={10861--10867},
  year={2025},
  organization={IEEE}
}

@article{dalal2024neural,
  title={Neural mp: A generalist neural motion planner},
  author={Dalal, Murtaza and Yang, Jiahui and Mendonca, Russell and Khaky, Youssef and Salakhutdinov, Ruslan and Pathak, Deepak},
  journal={arXiv preprint arXiv:2409.05864},
  year={2024}
}

@book{de2008computational,
  title={Computational geometry: algorithms and applications},
  author={De Berg, Mark and Cheong, Otfried and Van Kreveld, Marc and Overmars, Mark},
  year={2008},
  publisher={Springer}
}

@book{piegl2012nurbs,
  title={The NURBS book},
  author={Piegl, Les and Tiller, Wayne},
  year={2012},
  publisher={Springer Science \& Business Media}
}

@article{savitzky1964smoothing,
  title={Smoothing and differentiation of data by simplified least squares procedures.},
  author={Savitzky, Abraham and Golay, Marcel JE},
  journal={Analytical chemistry},
  volume={36},
  number={8},
  pages={1627--1639},
  year={1964},
  publisher={ACS Publications}
}

@inproceedings{lakshmipathy2021contact,
  title={Contact tracing: A low cost reconstruction framework for surface contact interpolation},
  author={Lakshmipathy, Arjun and Bauer, Dominik and Pollard, Nancy S},
  booktitle={2021 IEEE/RSJ International Conference on Intelligent Robots and Systems},
  pages={5165--5172},
  year={2021},
  organization={IEEE}
}

@article{guzey2025dexterity,
  title={Dexterity from Smart Lenses: Multi-Fingered Robot Manipulation with In-the-Wild Human Demonstrations},
  author={Guzey, I. and Qi, H. and Urain, J. and Wang, C. and Yin, J. and Bodduluri, K. and Lambeta, M. and Pinto, L. and Rai, A. and others},
  journal={arXiv preprint arXiv:2511.16661},
  year={2025}
}

@article{yin2025osmo,
  title={OSMO: Open-Source Tactile Glove for Human-to-Robot Skill Transfer},
  author={Yin, J. and Qi, H. and Wi, Y. and Kundu, S. and Lambeta, M. and Yang, W. and Wang, C. and Wu, T. and Malik, J. and Hellebrekers, T.},
  journal={arXiv preprint arXiv:2512.08920},
  year={2025}
}

@article{mao2025visuo,
  title={Visuo-Acoustic Hand Pose and Contact Estimation},
  author={Mao, Y. and Yoo, U. and Yao, Y. and Syed, S. N. and Bondi, L. and Francis, J. and Oh, J. and Ichnowski, J.},
  journal={arXiv preprint arXiv:2508.00852}
}

@inproceedings{Handa2019DexPilotVT,
    title={DexPilot: Vision-Based Teleoperation of Dexterous Robotic Hand-Arm System},
    author={A. Handa and K. V. Wyk and W. Yang and J. Liang and Y. W. Chao and Q. Wan and S. Birchfield and N. D. Ratliff and D. Fox},
    booktitle={IEEE International Conference on Robotics and Automation},
    pages={9164--9170},
    year={2019}
}

@INPROCEEDINGS{sivakumar2022telekinesis,
    title={Robotic Telekinesis: Learning a Robotic Hand Imitator by Watching Humans on Youtube},
    author={Sivakumar, A. and Shaw, K. and Pathak, D.},
    booktitle = {Robotics: Science and Systems},
    year={2022}
}

@article{naughton2024respilot,
  title={Respilot: Teleoperated finger gaiting via gaussian process residual learning},
  author={Naughton, P. and Cui, J. and Patel, K. and Iba, S.},
  journal={arXiv preprint arXiv:2409.09140},
  year={2024}
}

@inproceedings{yin2025geometric,
  title={Geometric retargeting: A principled, ultrafast neural hand retargeting algorithm},
  author={Yin, Z. H. and Wang, C. and Pineda, L. and Bodduluri, K. and Wu, T. and Abbeel, P. and Mukadam, M.},
  booktitle={IEEE/RSJ International Conference on Intelligent Robots and Systems},
  pages={17376--17382},
  year={2025}
}

@article{kostrikov2020image,
  title={Image augmentation is all you need: Regularizing deep reinforcement learning from pixels},
  author={Kostrikov, I. and Yarats, D. and Fergus, R.},
  journal={arXiv preprint arXiv:2004.13649},
  year={2020}
}

@article{lakshmipathy2023contactedit,
    title={Contact Edit: Artist Tools for Intuitive Modeling of Hand-Object Interactions}, 
    author={A. S. Lakshmipathy and N. Feng and Y. X. Lee and M. Mahler and N. S. Pollard},
    volume = {42},
    number = {4},
    articleno = {45},
    year = {2023},
    journal = {ACM Transactions on Graphics}
}

@article{mandi2025dexmachina,
  title={Dexmachina: Functional retargeting for bimanual dexterous manipulation},
  author={Mandi, A. and Hou, Y. and Fox, D. and Narang, Y. and Mandlekar, A. and Song, S.},
  journal={arXiv preprint arXiv:2505.24853},
  year={2025}
}

@article{pan2025spider,
  title={Spider: Scalable physics-informed dexterous retargeting},
  author={Pan, C. and Wang, C. and Qi, H. and Liu, Z. and Bharadhwaj, H. and Sharma, A. and Wu, T. and Shi, G. and Malik, J. and Hogan, F.},
  journal={arXiv preprint arXiv:2511.09484},
  year={2025}
}

@article{delpreto2022actionsense,
    title={Actionsense: A multimodal dataset and recording framework for human activities using wearable sensors in a kitchen environment},
    author={DelPreto, J. and Liu, C. and Luo, Y. and Foshey, M. and Li, Y. and Torralba, A. and Matusik, W. and Rus, D.},
    journal={Advances in Neural Information Processing Systems},
    volume={35},
    pages={13800--13813},
    year={2022}
}

@article{song2025opentouch,
    title={OPENTOUCH: Bringing Full-Hand Touch to Real-World Interaction},
    author={Song, Y. R. and Li, J. and Fu, R. and Murphy, D. and Zhou, K. and Shiv, R. and Li, Y. and Xiong, H. and Owens, C. E. and Du, Y. and others},
    journal={arXiv preprint arXiv:2512.16842},
    year={2025}
}

@incollection{paszke2019pytorch,
    title = {PyTorch: An Imperative Style, High-Performance Deep Learning Library},
    author = {Paszke, Adam and Gross, Sam and Massa, Francisco and Lerer, Adam and Bradbury, James and Chanan, Gregory and Killeen, Trevor and Lin, Zeming and Gimelshein, Natalia and Antiga, Luca and Desmaison, Alban and Kopf, Andreas and Yang, Edward and DeVito, Zachary and Raison, Martin and Tejani, Alykhan and Chilamkurthy, Sasank and Steiner, Benoit and Fang, Lu and Bai, Junjie and Chintala, Soumith},
    booktitle = {Advances in Neural Information Processing Systems},
    pages = {8024--8035},
    year = {2019}
}

\end{document}